%% file: cal_eval.tex
\documentclass{article} 
\usepackage{iclr2022_conference,times}
\iclrfinalcopy
\input{math_commands.tex}

\usepackage{hyperref}
\usepackage{url}
\usepackage{wrapfig}

\usepackage{microtype}
\usepackage{graphicx}
\usepackage{subfigure}
\usepackage{booktabs} 
\usepackage{fancyhdr}
\usepackage{amsmath}
\usepackage{amssymb}
\usepackage{mathtools}
\usepackage{amsthm}

\usepackage{amsfonts}
\usepackage{bbm}

\usepackage[capitalize,noabbrev]{cleveref}

\theoremstyle{plain}

\theoremstyle{definition}

\theoremstyle{remark}

\usepackage[textsize=tiny]{todonotes}

\title{What is Your Metric Telling You? Evaluating Classifier Calibration under Context-Specific Definitions of Reliability}

\author{John Kirchenbauer \\
Department of Computer Science \\
University of Maryland, College Park\\
\texttt{jkirchen@umd.edu} \\
\And
Jacob Oaks \& Eric Heim \\
Software Engineering Institute \\
Carnegie Mellon University\\
\texttt{\{jroaks, etheim\}@sei.cmu.edu} \\
}

\begin{document}

\maketitle

\begin{abstract}
Classifier calibration
has received recent attention from the machine learning community due both to its practical utility in facilitating decision making, as well as the observation that modern neural network classifiers are poorly calibrated.
Much of this focus has been 
towards the goal of learning classifiers such that 
their output with largest magnitude (the ``predicted class'') is calibrated.
However, this narrow interpretation of classifier 
outputs 
does not adequately capture the variety of practical use cases in which classifiers can 
aid in decision making.
In this work, we argue that more expressive metrics must be developed that accurately measure calibration error for the specific context in which a classifier will be deployed.
To this end, we derive a number of different metrics using a generalization of Expected Calibration Error (ECE) that measure calibration error under different definitions of \emph{reliability}.
We then provide an extensive empirical evaluation of commonly used neural network architectures and calibration techniques 
with respect to 
these metrics.
We find that: 1) definitions of ECE that focus solely on the predicted class fail to accurately measure calibration error under a selection of practically useful definitions of reliability and 2) many common calibration techniques fail to improve calibration performance uniformly across ECE metrics derived from these diverse definitions of reliability.

\end{abstract}

\section{Introduction}
Classification is arguably the most widely-studied problem in machine learning with applications ranging from medical diagnosis to visual object identification in autonomous vehicles.
Despite the variety of classification tasks, the definitions of common evaluation metrics are often based 
on the assumption that model outputs will be used in specific ways (i.e. a particular decision rule), and such assumptions affect how a classifier's ``predictive'' performance is measured.
In practice, classifier outputs are often consumed by either a human (such as for diagnosis in medical care \citep{jiang2012calibrating}) or higher-level reasoning algorithms (such as for pedestrian detection in autonomous vehicles \citep{yurtsever2020survey}).
In these settings, classifiers are called upon to inform 
complex and task-specific decision processes. 
Learning a classifier that can express not just class ``predictions'', but \emph{confidence} associated with classes, is one way to better inform downstream decision making.
For example,
a clinician can order additional lab tests to
clarify their diagnosis 
if a classifier
strongly predicts an unexpected condition. 
Similarly, a self-driving system can defer to higher resolution sensors if a vision based classifier is uncertain whether an object in the road is a pedestrian.

Both because of the practical utility in estimating classifier confidence, and the observation that modern neural network classifiers often fail to do so accurately \citep{guo2017calibration}, there has been a recent focus on \emph{classifier calibration} --  the problem of learning probabilistic classifiers whose confidence outputs match the empirical distribution of observed class labels. 
This interpretation of class probabilities enables the formal 
analysis of how well a classifier models true distributions over events.
Because of this, the term calibration is often used interchangeably with \emph{reliability}~\citep{brocker2008some}.
While this focus has revitalized the study of classifier calibration, metrics used to evaluate calibration techniques, such as the popular Expected Calibration Error (ECE)~\citep{naeini2015obtaining}, presuppose a specific interpretation of classifier outputs.
Most notably, it has become common to evaluate classifier calibration using metrics that measure error for only the most probable classifier output.
While such an evaluation may be adequate in certain cases, when more than a single output is used in decision making, metrics derived from broader definitions of reliability may be required.

We argue that in order to meaningfully evaluate calibration, there must be a focus on the \emph{context} in which a classifier is deployed.
More specifically, the practical assessment of calibration requires consideration for 
isolated cases that are of interest in the particular application of the classifier (such as high-risk scenarios), and 
how classifier outputs are consumed.
This necessitates metrics that map to a variety of definitions of reliability for the purpose of context-specific evaluation of classifier calibration. 

Recognizing this need, this work contributes the following:
\begin{enumerate}
    \item We reframe the generalized form of the expected calibration error introduced in~\cite{vaicenavicius2019evaluating} 
    as a set of modular components.
    \item Within this framework, we discuss how common components from prior work map to classifier use cases and introduce novel ones that result in metrics that more specifically target other use cases.
    \item 
    We extensively evaluate
    a survey of classifier calibration techniques and find that many 
    perform inconsistently when evaluated against the diverse set of ECE metrics derived from different definitions of reliability.
\end{enumerate}

Through our evaluation we find that 1) relative performance as measured by traditional ECE metrics does not correspond to relative performance as measured by different context-specific metrics 2) many common calibration methods that reduce traditional ECE over baselines do not uniformly reduce 
variants of the generalized ECE metric
better suited for different contexts.
We believe this provides strong motivation for more focused calibration evaluation.
In addition, we believe that by posing ECE metrics as modular measures that can be adapted, we provide practitioners with the tools they need to properly evaluate their classification models for the context in which they will be used.

\section{Related Work}
Much of the recent focus on classifier calibration from the machine learning community can be attributed to the observation that modern deep neural network classifiers are \emph{overconfident}.  
Namely, \citet{guo2017calibration} provided an empirical evaluation showing that common network architectures trained using standard training procedures produce models that place an incorrectly high amount of probability on a single class, and that simple techniques can often reduce this effect.
As such, the authors used a form of ECE in their evaluation that measures the miscalibration of the most probable class output by the classifier, 
a choice mirrored in a number of subsequent  works~\citep{thulasidasan2019mixup,muller2019when,mukhoti2020calibrating,alexandari2020maximum}.
In the multi-class setting, such a metric makes practical sense when only the confidence of the most probable class (i.e. the ``prediction'') is considered in downstream reasoning.
However, a probabilistic classifier's output is a distribution over all classes and can be interpreted in a number of ways not captured by this ``Top-1'' ECE.
In this work, we investigate ECE metrics that assume different definitions of reliability and deployment contexts, and show how such assumptions affect classifier evaluation.

Two recent works,
~\citet{ovadia2019can} and~\citet{nixon2019measuring}, also sample the state of the art in calibration and evaluate them on a variety of data sets and neural network architectures.
The former focuses on evaluating techniques under data set shift, which we do not consider in this work.
The latter focuses on addressing issues with computing the ECE metric itself, while our  work focuses on expanding ECE to different contexts.


A number of alternatives to ECE have been used to measure calibration performance of classifiers. 
Brier score~\citep{brier1950verification} and negative log likelihood~\citep{lakshminarayanan2017simple} have desirable properties associated with being proper scoring rules, but do not directly capture the concept of calibration error.
Other metrics such as Maximum Mean Calibration Error~\citep{widmann2019calibration} 
and Kernel Calibration Error~\citep{kumar2018trainable} 
 are more explicit in defining calibration targets to measure error.
 These  were largely developed as alternatives to ECE to address the difficulty of estimating expectation using histogram binning.
We discuss this issue in Section~\ref{sec:HistEst} of the appendix, and provide some practical intuition on how to alleviate it.

We adopt the Generalized Expected Calibration Error (GECE) framework proposed in~\citet{vaicenavicius2019evaluating}.
Here, the authors propose a theoretical framework that introduces the concept of a \emph{calibration lens} as a means to formally define the relationship between a reliability condition and an ECE metric.
While they also 
sketch out the formalism for a few potentially useful lenses, such as the ``Top-$k$'' and ``Grouping'' lenses discussed in Section~\ref{sec:Lenses}, they mainly use the concept of a lens to relate a ``canonical'' reliability condition to the Top-1 reliability condition.
In this work we use the GECE framework as a means to realize a variety of methods for evaluating classifier calibration.
In Section~\ref{subsec:ECEFramework} we reframe GECE into a set of modular components, which we use to introduce a number of distance functions and selection operators that map to practical classifier use cases.
As such, this work can be seen as a natural extension of the original GECE work to evaluate the practical considerations and effects of using GECE to derive metrics for context-specific calibration evaluation.

\section{Preliminaries}\label{sec:prelims}
\subsection{Reliability Conditions}
Let $\mathcal{X}$ be a domain of instances and $\mathcal{Y}$ be a set of $k$ classes.
Let $p\left(X\in\mathcal{X},Y\in\mathcal{Y}\right)$ be a joint distribution over instances and classes, and similarly let $p\left(Y|X\right) = p\left(X,Y\right)/p\left(X\right)$ be a conditional distribution over classes given an instance.
The goal of \emph{probabilistic classification} is to find a classifier $g : \mathcal{X} \mapsto \Delta^{k-1}$ that estimates $p\left(Y|X\right)$ given a set of finite samples $\mathcal{D}_{train} = \left\{\left(x_1,y_1\right),...,\left(x_n,y_n\right)\right\} \mathtt{\sim}\ p\left(X,Y\right)$.
Here, we assume for concreteness that $\hat{g}$ outputs the predicted distribution over classes in the form of vectors in
the probability simplex $\Delta^{k-1} = \left\{\mathbf{s} \in [0,1]^{k} : \sum_{i=1}^k \mathbf{s}_i = 1\right\}$, and labels $y$ are represented by one-hot encodings $\mathbf{y}$, except in the binary ($k=2$) case where $g(x)\in[0,1]$ and $\mathbf{y} = y \in \{0,1\}$. 
In many cases, it is impossible to learn a classifier that recovers $p\left(Y|X\right)$ exactly.
However, even though learned models imperfectly estimate this distribution, it is still 
desireable that
classifier outputs can be interpreted as accurate estimates of \emph{confidence} in an instance's class. 


One way to formalize this is through a \emph{calibration \emph{or} reliability condition}~\citep{murphy1987general}; the most general of which is the \emph{Canonical} or \emph{Full} reliability condition:
\begin{equation}
\mathbb{P}\left(Y | g\left(x\right)\right) = g\left(x\right) 
\label{eq:canon_cond}
\end{equation}
This condition specifies that the distribution over classes, conditioned on the classifier's output, is equal to the classifier's output.
For example, for all $\left(x,y\right) \mathtt{\sim}\  P(X,Y)$ in which $g(x)$ outputs $[0.7,0.2,0.1]$, $70\%$ of the time $y$ is class $1$,  $20\%$ of the time it is class $2$, and $10\%$ of the time it is class $3$.
For problems with many classes, satisfying the full reliability condition is often unnecessary to facilitate practical classifier use cases.
As a result, often weaker conditions are considered, such as the \emph{Top-1} condition:
\begin{equation}
\mathbb{P}\left(Y = \argmax_y{g(X)} | \max{g\left(X\right)}\right) = \max{g\left(X\right)}
\label{eq:max_cond}
\end{equation}
Here, only the maximum output of the classifier is considered.
A perfectly calibrated classifier under this condition would predict the correct class by the max output decision rule 70\% of the time when it's maximum output is 0.7.
In general, a classifier that satisfies one reliability condition does not imply that it satisfies another.
A key relationship to note is that satisfaction of the Full condition,~\eqref{eq:canon_cond}, implies the same for the Top-1 condition,~\eqref{eq:max_cond}~\citep{vaicenavicius2019evaluating}, but the converse does not necessarily hold true.


\subsection{From Conditions to Lenses to Error}
\label{subsec:ECEFramework}
While reliability conditions provide a principled way to specify when a classifier is perfectly calibrated, classifiers are rarely able to satisfy non-trivial conditions.
As a result, it is useful to specify a way of measuring the discrepancy between classifier outputs and the ideal.
For this, we adopt the generalized ECE (GECE) framework first presented in~\citet{vaicenavicius2019evaluating}.
Here, we will present it with different notation in order to modularize the framework into components that can be designed to empirically quantify calibration performance for different contexts.
These components include: A \emph{lens} that expresses how to transform both classifier outputs and ground truth labels into forms that match a calibration condition, a \emph{distance function}
that can be used to measure the difference between lensed outputs and labels, and an \emph{estimation scheme} to estimate the error over a domain.

A lens itself has two components.
First, an \emph{output transformation function}, defined as the function $o : \Delta^{k-1} \mapsto \Delta^{k'-1}$, maps a classifiers output $g(x) = \mathbf{g}$ to induced outputs ${\mathbf{g}}'$.
Second, a \emph{target transformation function}, defined as the function $t : \Delta^{k-1} \mapsto \Delta^{k'-1}$, maps labels $\mathbf{y}$ to induced labels $\mathbf{y}'$.
These two functions define an \emph{induced classification problem} that implies a particular calibration condition. 
To measure the degree in which a classifier violates a calibration condition, we require a \emph{distance function} $d: \Delta^{k'-1} \times \Delta^{k'-1} \mapsto \mathbb{R}$ that compares classifier outputs to labels and produces a single scalar measurement of \emph{error}.
In most cases this is a proper distance metric such as total variation distance (TVD) $d\left(\mathbf{g}',\mathbf{y}'\right) = \frac{1}{2}||\mathbf{g}'-\mathbf{y}'||_1$.

To measure error over the entirety of the instance and class domains, we take expectation of the error over $P(X,Y)$. 
In the case where there are only finite samples for evaluation, we must estimate the expectation with an \emph{estimation scheme}.
One of the most popular estimation schemes is \emph{histogram estimation} that partitions $\Delta^{k'-1}$ into a finite number of bins and estimates the expectation as a sum of errors between expected outputs and targets in each bin.
Let $\mathbb{B} = \left\{\mathcal{B}_1,...,\mathcal{B}_b\right\}$ be a partitioning of $\Delta^{k'-1}$, and $|\mathbb{B}| = \sum_{\mathcal{B}\in\mathbb{B}}|\mathcal{B}|$.
A sample $x$ is ``binned'' in $\mathcal{B}$ if $o(g(x)) \in \mathcal{B}$.
Let $\bar{\mathbf{g}}_{\mathcal{B}}$ be the mean lensed outputs $o(g(x)) = \mathbf{g}'$ for all $x$ binned in $\mathcal{B}$.
Similarly, let $\bar{\mathbf{y}}_{\mathcal{B}}$ be the mean lensed labels $t(\mathbf{y})=\mathbf{y}'$ for all $x$ binned in $\mathcal{B}$.
The histogram estimation of the expectation of calibration error is defined as:
\begin{equation}\label{eq:ECEhist}
    GECE(g) = \sum_{\mathcal{B}\in\mathbb{B}} \frac{|\mathcal{B}|}{|\mathbb{B}|}d\left(\bar{\mathbf{g}}_{\mathcal{B}},\bar{\mathbf{y}}_{\mathcal{B}}\right)
\end{equation}

As an example, the Top-1 or ``Traditional'' ECE metric used in~\citet{guo2017calibration} is recovered in this framing of ECE by the following: 
 \begin{itemize}
     \item $o(g(x)) = \max g(x)$
     \item $t(\mathbf{y}) = \mathbf{y}[\argmax g(x)]$
     \item $d\left(\bar{\mathbf{g}},\bar{\mathbf{y}}\right) = \frac{1}{2}||\bar{\mathbf{g}}-\bar{\mathbf{y}}||_1$
     \item Histogram binning -- 15 uniform bins over [0,1]
 \end{itemize}



\section{Context-Specific Calibration Evaluation}
The generalized ECE framework allows for the design of metrics that reflect how classifiers are used in practice. This is accomplished by choosing a lens derived from a reliability condition, and a distance function that defines how error is calculated.
In this section we discuss a number of lenses and distance functions that can be used to map metrics to practical use cases.
In addition, we discuss the concept of a selection operator as a means to isolate specific instances for evaluation.
We leave a discussion of estimation schemes (histogram estimation) to Section~\ref{sec:HistEst} of the appendix.

\subsection{Selection Operators}
In order to formalize the common practice of selecting subsets of instances for evaluation that map to important scenarios, let $\lambda$ be a \emph{selection operator} such that $\lambda(\mathcal{D}_{test}) \subseteq \mathcal{D}_{test}$ that selects sample pairs $(x,\mathbf{y})$ for evaluation based on conditions over labels or classifier outputs. 

\textbf{Label Conditions}
The label conditional selection operator $\lambda_{y=c}$ selects instance/label pairs from $\mathcal{D}_{test}$ with label $c$.
This is useful when particular events are of interest.
For instance, when evaluating a classifier for an autonomous vehicle it is often important to evaluate how the model performs in high risk cases such as in instances labeled ``pedestrian'' (i.e. $\lambda_{y=\text{``ped''}}$).
This is analogous to classifier predictive metrics that condition on labels, an example being the true positive rate often used in binary settings.

\textbf{Output Conditions}
Similarly, the output conditional selection operator $\lambda_{g(x)\diamond z}$ selects instance/label pairs based on the output of $g$.\footnote{Note here we use $\diamond$ to represent an abstract conditional operator, as there are a number of conditions that can be specified.}
For instance, consider the case of a binary classifier that determines whether a patient has a disease.
It may be useful to evaluate the classifier in the cases where it is very confident (e.g. $\lambda_{g(x)>0.95}$), as such a condition can be used as the basis for a system that alerts a physician.
Output conditions are analogous to classifier predictive metrics that condition on outputs such as positive predictive value.

\subsection{Lenses}
\label{sec:Lenses}

While the Full calibration condition is often sufficient to capture desired definitions of reliability, 
it is difficult to learn a model that satisfies it.
Alternatively, the Top-1 condition is much less restrictive, but may not sufficiently represent how a classifier is used.
In this section, we discuss two lenses derived from different reliability conditions than the Full or Top-1, and argue for their practical utility.

\textbf{Top-$k$}
Consider the case where a human is called upon to make decisions based in part on classifier confidence, such as a physician being assisted with a diagnosis.
Here, the user may want to consider more than just the most probable class, but not all classes
(e.g. the top 5 most probable ailments for a patient).
This motivates the \emph{Top-$k$} lens, where:
\begin{equation}
o(g(x)) = \max_k g(x), t(\mathbf{y}) = \mathbf{y}\left[\argmax_k{g(x)}\right] 
\label{eq:TopKLens}
\end{equation}
This generalizes the Top-1 lens by allowing for selection of outputs and targets based on the highest $k$ values in $g(x)$.
This is similar in spirit to the commonly used top-$k$ accuracy metric used for measuring predictive performance.

\textbf{Grouping}
In certain applications, classes fall into natural groupings that are each handled differently.
Consider a social media platform that uses a classifier to flag content that violates its terms of use, where 
the disciplinary action for within subsets of violations is the same.
This represents a case where miscalibration within a group does not affect downstream decision making.
This case and others motivates metrics that measure calibration error with respect to groupings for which different downstream actions are taken.
For this we propose the $grouping$ lens.
Let $\mathbb{G} = \left\{\mathcal{G}_1,\mathcal{G}_2,...\right\}$ be a grouping of $\mathcal{Y}$, such that $\forall_{i,j} \mathcal{G}_i \cap \mathcal{G}_j = \emptyset$ and $\bigcup_{\mathcal{G}\in\mathbb{G}} \mathcal{G} = \mathcal{Y}$.
Let the \emph{grouping operator} for $\mathbb{G}$ be defined as: 

\vspace{-0.4cm}
\begin{equation}
    \delta_{\mathbb{G}}\left(\mathbf{z}\right) = \left[\sum_{i\in\mathcal{G}_1}\mathbf{z}[i],...,\sum_{j\in\mathcal{G}_{|\mathbb{G}|}}\mathbf{z}[j]\right]
\end{equation}

The grouping lens for operator $\delta_{\mathbb{G}}$ is defined as:
\begin{equation}
o(g(x)) = \delta_{\mathbb{G}}(g(x)), t(\mathbf{y}) = \delta_{\mathbb{G}}(\mathbf{y}) 
\label{eq:GroupingLens}
\end{equation}
Here, we transform the probability space over classes ($\Delta^{k-1}$) to the probability over groups ($\Delta^{|\mathbb{G}|-1})$
by summing the outputs corresponding to the classes within each group.
As such, a grouping is a logical disjunction of the events of class membership of the classes within the group.


\subsection{Distance Functions}
While distance functions based on $\ell_1$ and $\ell_2$ norms are most commonly used for ECE~\citep{nixon2019measuring},
in this section, we propose a non-metric distance function as means to better 
quantify error specifically when classifier outputs are directly consumed by a human at decision time in the binary classification setting (i.e. $\mathbf{y} = y$ and $\bar{\mathbf{y}} = \bar{y}$).

\textbf{Inter-Interval Distance}
It has been shown that humans are poor at reasoning about continuous values~\citep{teodorescu2016absolutely} such as probabilities, especially at small scales~\citep{resnick2017dealing}.
As such, when designing classifiers that will aid human decision making, it is unreasonable to expect them to understand the exact, continuous confidence values that are output.

In the human factors community, it is common to utilize a Likert scale~\citep{likert1932technique}, a set of ordinal categories representing ranges of values over an interval, to provide more coarse estimates of continuous values.
Indeed, Likert scales have been used both as a means to elicit~\citep{xue2017active}, and visualize~\citep{guo2019visualizing}, confidence information.
%

In order to evaluate a classifier's ability to express confidences that reflect Likert scales, we require metrics that reflect the error tolerance specified by Likert categories.
For this, we utilize the following non-metric distance function:
\begin{equation}
d_{[l,h]}\left(\mathbf{g},\bar{y}\right) = \max(0,\max\left(l-\bar{y},\bar{y}-h\right))
\label{eq:Inter-IntervalDistance}
\end{equation}
When the Inter-Interval Distance~\eqref{eq:Inter-IntervalDistance} is used for the expected error calculation as defined in~\eqref{eq:ECEhist}, error in each Likert category is only incurred for a bin when its mean label exceeds the bounds defined by the category's interval $\left[l,h\right]$.
If this distance function is used in tandem with the output selection operators that select only instances where classifier outputs fall within the respective interval $\left[l,h\right]$ (i.e. $\lambda_{g(x)\geq l}$ and $\lambda_{g(x)\leq h}$), classifier calibration can be evaluated with respect to a Likert category defined by $\left[l,h\right]$.
As an example, if the interval [0,0.2] 
denotes a ``very low confidence'' category, through the use of~\eqref{eq:Inter-IntervalDistance} and corresponding output selection operators, the degree to which a classifier reliably outputs values in the 
low confidence regime can be measured.

\section{Empirical Evaluation}\label{sec:empiricalEval}
Equipped with the means to perform context-focused evaluation of calibration performance, we wish to determine whether recently proposed classifier calibration techniques that have been shown to reduce Top-1 ECE also can reduce other variants within the generalized ECE framework.
Further, we wish to highlight the broader utility of generalizing the ECE metric family by providing a number of case studies in calibration evaluation that reflect a number of potential use cases of probabilistic classifiers.
To these ends, we performed an empirical evaluation that focuses on the task of calibrating image classifiers
that spans a number of neural network architectures, data sets, metrics, and calibration techniques.
Due to page limits, this section reports only on a subset of the full set of experiments performed that best represent our findings.
Experiments not described in this section can be found in Section~\ref{sec:AddExpRes}.


In evaluating techniques used to calibrate neural networks, we choose to view techniques as \emph{interventions} on a base model, i.e a calibration technique is an explicit augmentation of a neural network or its training procedure.
\textit{Training-time} interventions either augment training data, or define an  objective to bias the parameters of a neural network towards values that lead to a more calibrated classifier, while \textit{post-processing} interventions transform a model's output.
Training-time interventions in our experiments include: Label smoothing (LS)~\citep{szegedy2016rethinking,muller2019when}, Mix-up (MU)~\citep{zhang2018mixup,thulasidasan2019mixup}, and Focal loss (FL)~\citep{lin2017focal,mukhoti2020calibrating}.
Post-processing interventions in our experiments include: Temperature scaling (TS)~\citep{jaynes1957information,guo2017calibration}, Bias-corrected temperature scaling (BCTS)~\citep{alexandari2020maximum}, and Histogram binning (HB)~\citep{zadrozny2001obtaining}.
While we leave full descriptions of these techniques to their respective works, each of these techniques have been shown to reduce the overconfidence effect reported in~\citep{guo2017calibration}.
Label smoothing, focal loss, and temperature scaling address overconfidence by pushing classifier outputs towards the uniform distribution, while the others are more opaque in their specific effect on the predictive distribution. 

\begin{figure*}[t]
    \centering
    \includegraphics[width=\textwidth]{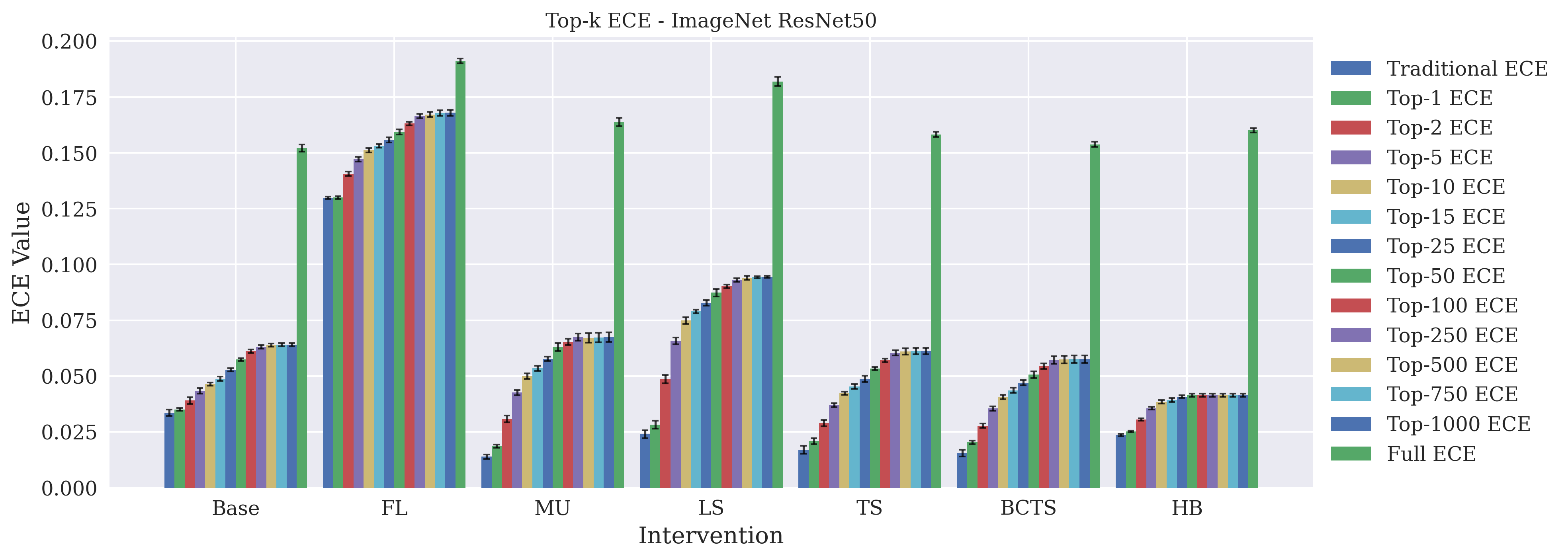}
    \caption{Top-$k$ ECE (All Interventions, ResNet50, ImageNet)}
    \label{fig:TopKECEImageNet}
\end{figure*}
%


Unless otherwise stated, for all experiments we train a 50-layer residual network (ResNet50) as the base model, use TVD as a distance function, run 5 trials, and show the means and standard deviations as error bars in the accompanying figures.
Further, we adopt a dynamic binning strategy with free parameters set using the heuristic discussed in Section~\ref{sec:HistEst}.
We display Top-1 or ``Traditional'' ECE, as defined at the end of Section~\ref{sec:prelims} and clarified in Section~\ref{sec:topKResults}, alongside other proposed metrics to provide a comparison with how classifier calibration is most commonly measured in other works.
To train models for our experiments we first establish a baseline model by sweeping over relevant optimization hyperparameters to maximize top-1 accuracy on its respective validation set. 
Then, we add the calibration intervention component to the baseline model and tune any hyperparameters specific to the technique to optimize Traditional ECE without exceeding a 5\% decrease in accuracy. 
We do this with 5 different random seeds in our training procedures for the 5 trials. 
We will release our experiment code with all hyperparameters and seeds used upon publication.

\begin{wrapfigure}{r}{0.5\textwidth}
    \centering
    \vspace{-1cm}
    \includegraphics[width=\linewidth]{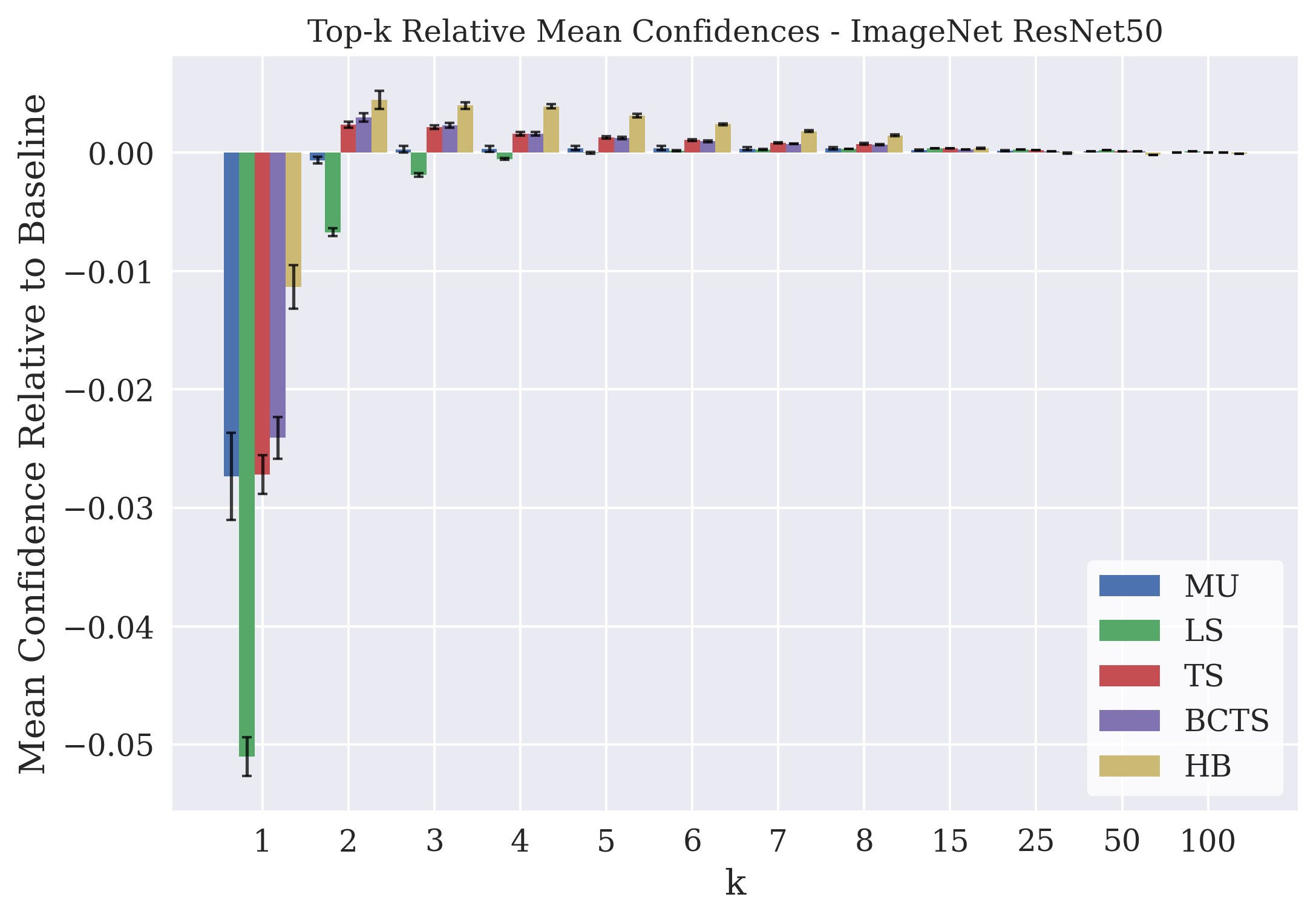}
    \vspace{-0.7cm}
    \caption{Mean Top-$k$ confidences, relative to baseline (All Interventions except Focal Loss, ResNet50, ImageNet)}
    \label{fig:TopKConfImageNet}
    \vspace{-0.5cm}
\end{wrapfigure}

%
\subsection{Top-$k$ Lens}\label{sec:topKResults} 

For our first experiment, we investigate how calibration techniques affect not just the most probable output of classifiers, but other less probable classes that could influence decision making in many contexts. Figure~\ref{fig:TopKECEImageNet} shows the top-$k$ ECE's for all interventions and the baseline ResNet50 on the popular ImageNet~\citep{deng2009imagenet} data set (baseline mean/std accuracy: 0.761/0.0013, 2 trials in this case). First, note the bars labeled ``Traditional ECE'' and ``Top-1 ECE''. The former, is the Top-1 lens ECE estimated with 15 uniform bins, and the latter is the same but with adaptively chosen bins. For most methods the two metrics score Top-1 calibration similarly, but in some cases, the binning strategies result in different values. 

Excluding focal loss, all interventions reduce the Traditional and Top-1 ECE over the baseline.\footnote{
As we are unaware of published results on improving ImageNet classification using Focal loss,
we used a sampling of the best performing settings on other data sets reported in~\citet{mukhoti2020calibrating}, but were unable to find a hyperparameter setting that was able to reduce Traditional ECE.
ImageNet was the only data set in our experiments where this was the case, and note that it is possible there exists a hyperpameter setting that results in better Top-1 ECE.}
However, at values of $k=5$ and higher, almost all interventions improve top-$k$ ECE values very little, if at all. Mix-up, Temperature scaling, and Bias corrected temperature scaling leave ECE values of $k$ greater than $5$, as well as Full ECE, at a similar value to the baseline, while Label smoothing performs worse than the baseline with respect to all ECE types other than Traditional and Top-1. Histogram Binning is the only technique that improves ECE at most values of $k$, though it still fails to improve ECE under the Full lens.

%
\begin{figure*}[t]
    \centering
    \includegraphics[width=\textwidth]{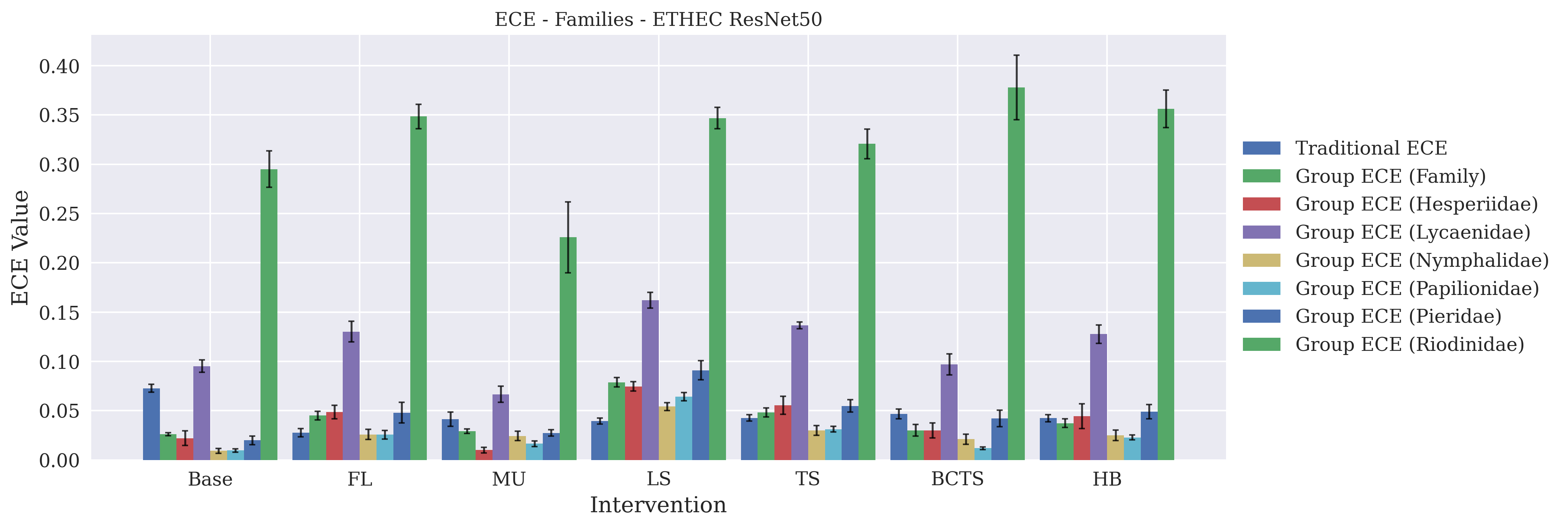}
    \caption{Group-wise ECEs, (All Interventions, ResNet50, ETHEC)}
    \label{fig:GroupingECE}
\end{figure*}

Figure~\ref{fig:TopKConfImageNet}, which shows the mean confidence values for the $k$th most probable class relative to the baseline for each intervention, provides some insight into these phenomena. Here we see that all calibration techniques reduce the most confident output of the base classifier, thus improving Top-1 ECE by reducing overconfidence. 
However, each intervention also affected the confidences of the other 999 classes differently. 
For instance, label smoothing reduced confidences for the 2nd through 5th most probable classes.
This had the effect of raising Top-$k$ ECE for $k \in [2,5]$ compared to the baseline.
This indicates that the base classifier was \emph{under-confident} for these classes, and label smoothing incorrectly reduced the corresponding confidences.

Histogram binning, on the other hand, raised the confidence of these classes and was able to reduce their corresponding Top-$k$ ECEs relative to the baseline.
In summary, 1) Traditional ECE is a poor indicator for performance in Top-$k$ (for $k>1$) and full ECE, and 2) different calibration techniques distribute probability mass differently, which determines how they perform when considering more of the 
output distribution than just the most probable class.

\begin{wrapfigure}{r}{0.5\textwidth}
    \centering
    \vspace{-0.8cm}
    \includegraphics[width=\linewidth,trim={0 0.3cm 0 0},clip]{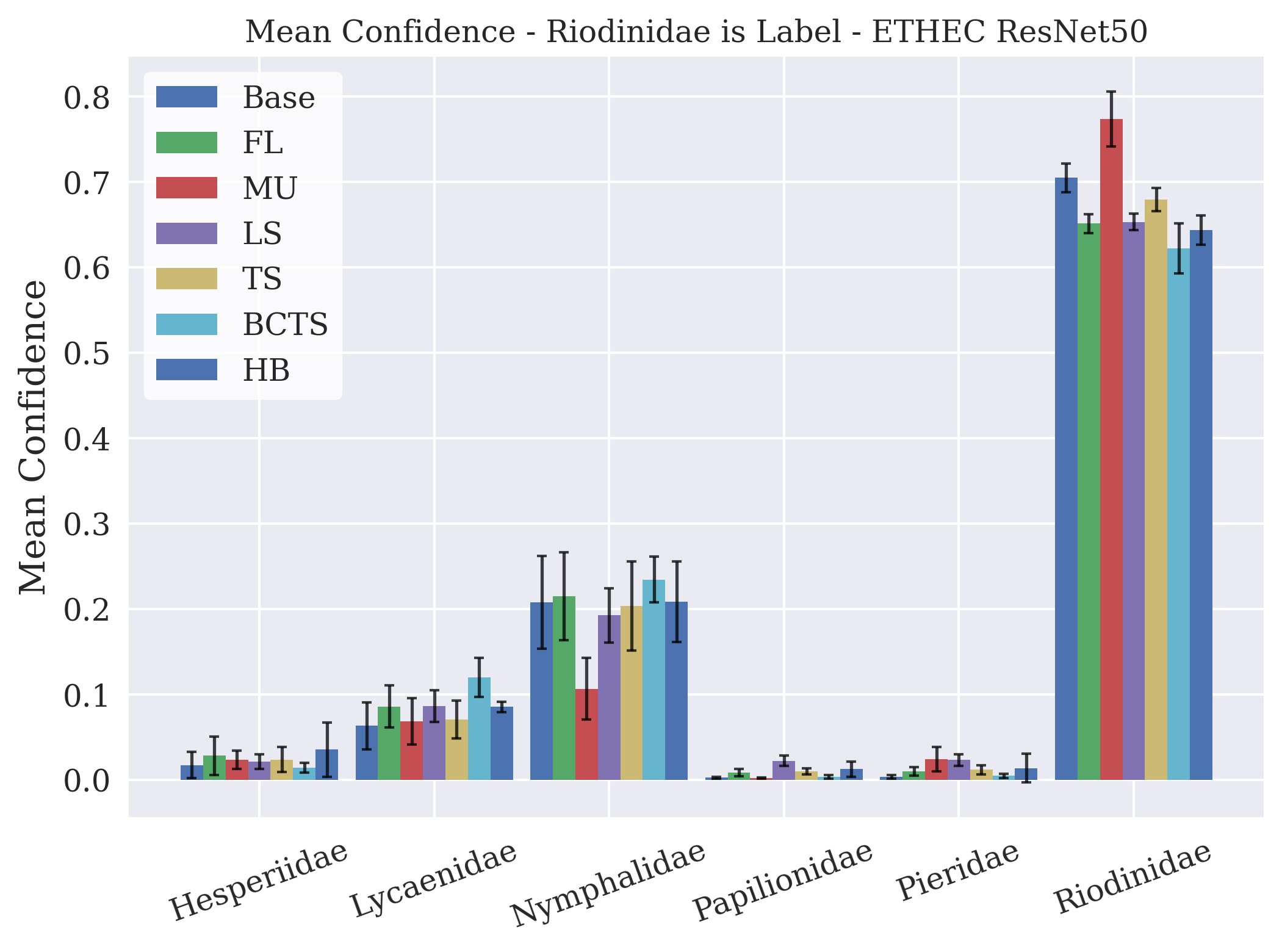}
    \vspace{-0.7cm}
    \caption{Mean family-wise confidence, conditioned on ``Rhiodinidae'' group (All Interventions, ResNet50, ETHEC)}
    \label{fig:GroupingConf}
    \vspace{-0.3cm}
\end{wrapfigure}

\subsection{Grouping Lens and Label Selection}\label{sec:groupResults}
To motivate the utility of calibration error measures that consider groups of classes, argued in Section~\ref{sec:Lenses},
we run an experiment using classifiers trained on the ETHEC data set~\citep{dhall2020hierarchical}.
This data set contains images of Lepidoptera and hierarchical taxonomic labels indicating the family, sub-family, genus, and species of each instance.
We train our model to classify the species of each image (baseline mean/std accuracy: 0.844/0.0049), but label sets higher in the hierarchy (such as family) represent natural groupings 
for evaluating classifier
calibration across groups.

Figure~\ref{fig:GroupingECE} shows family group-wise ECE metrics.
The first green bar for each intervention represents the the ECE after the family grouping lens is applied (inducing the six way family classification problem).
We see that while all interventions decrease Traditional ECE, some increase family-wise ECE 
suggesting that this improvement is achieved
by reducing probability on the most probable class
while
incorrectly increasing probability for classes across family groups.
The remaining bars show family group ECE after instances are selected using a group (label) conditional selection operator for each of the six families of Lepidoptera (inducing six, one-vs-five individual family classification problems).
This allows us to observe that instances belonging to two families (Lycanidae and Riodinidae) incur much higher family-wise ECE than the other families.


Figure~\ref{fig:GroupingConf} sheds some light on why these two errors are especially high.
Here, we plot the mean confidences for each class output by the models for only instances with the label Riodinidae (group selection operation).
We see that many of the techniques incorrectly reduce the confidence in the correct label over the baseline (besides Mix-up), which can explain the relative performance between techniques according to Riodinidae-conditioned group ECE shown in Figure~\ref{fig:GroupingECE}.
In addition, all models assign a significant amount of confidence in the Nymphalidae group, highlighting that the techniques tend to ``confuse'' Riodinidae instances as Nymphalidae to a higher degree than the other groups.
Interestingly, Lepidopterology literature suggests these two families are very closely related~\citep{brown2012chromosomal}.

%
\begin{figure*}
    \centering
    \includegraphics[width=\textwidth]{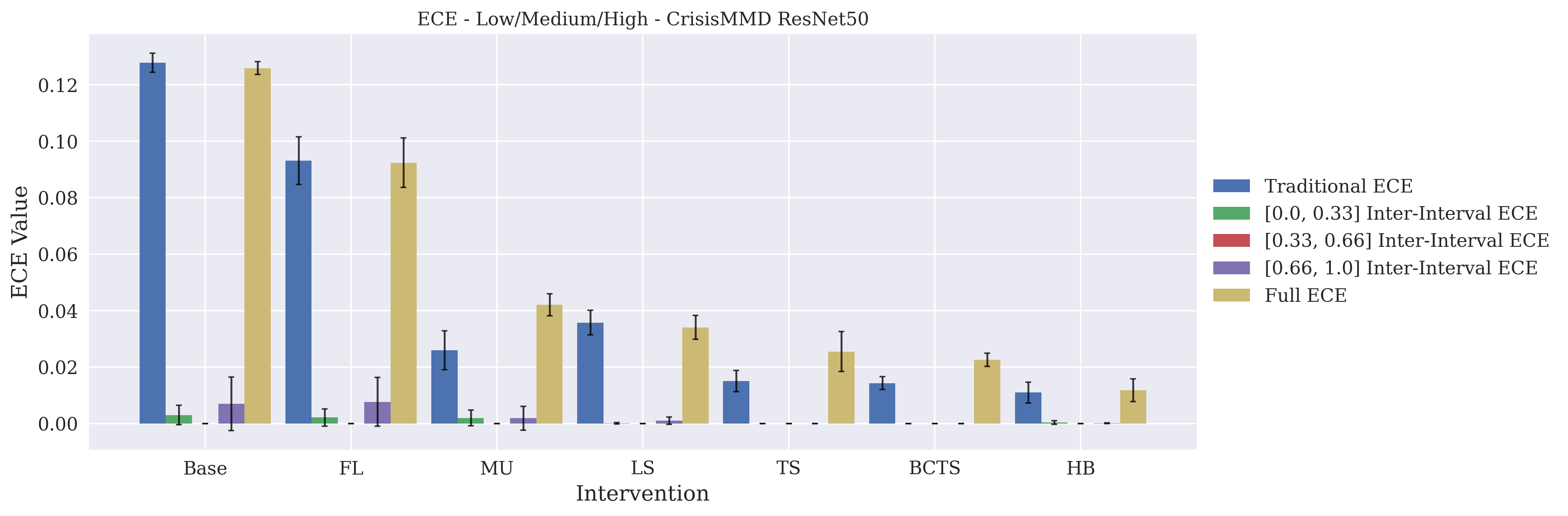}
    \caption{Interval-wise ECEs (All Interventions, ResNet50, CrisisMMD)}
    \label{fig:IntervalECE}
\end{figure*}
\begin{wrapfigure}{r}{0.5\textwidth}
  \centering
  \vspace{-0.5cm}
    \includegraphics[width=\linewidth,trim={0 0.3cm 0 0},clip]{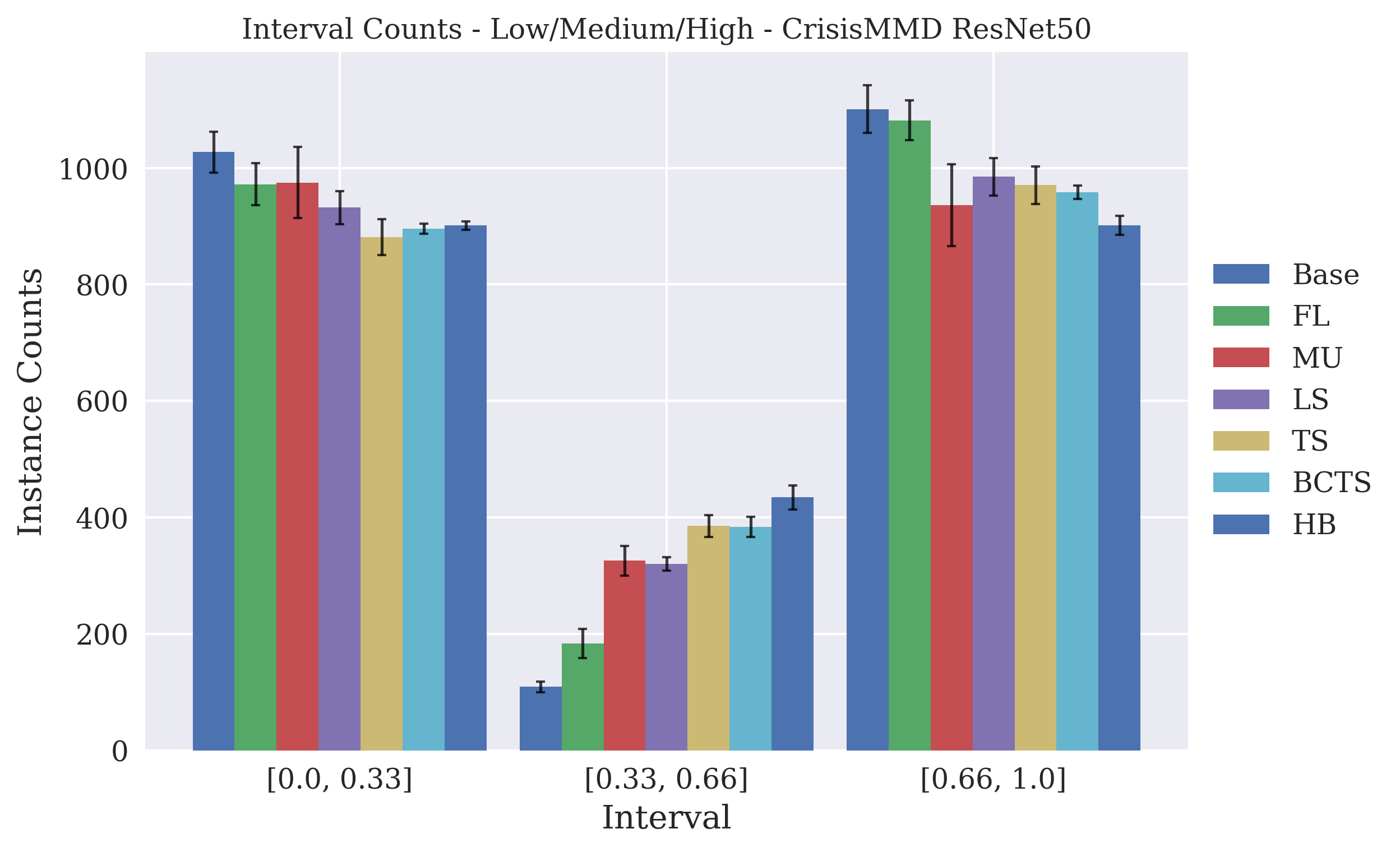}
    \vspace{-0.7cm}
    \caption{Test instances per interval (all interventions, ResNet50, CrisisMMD)}
    \label{fig:IntervalCount}
    \vspace{-0.5cm}
\end{wrapfigure}
\vspace{0.8cm}
\subsection{Inter-Interval Distance and Output Selection}\label{sec:interIntervalResults}
In our final experiment we highlight metrics for evaluating confidence under the assumption that it will be displayed to humans in a Likert scale.
For this we trained a classifier (baseline mean/std accuracy: 0.825/0.0023) on a version of the CrisisMMD data set~\citep{alam2018crisismmd} that consists of image tweets that were either labeled ``informative'' or ``not informative'' for analysis and tracking of humanitarian crises.
We evaluate classifier calibration for this problem under the assumption that classifier confidences will be used to express low ([0.0,0.33]), medium ([0.33,0.66]), and high ([0.66,1.0]) confidence categories to end-users.


Figure \ref{fig:IntervalECE} shows 
the Traditional and full ECE metrics as before but also includes three others that measure error for each category.
For instance, the green bar is achieved by applying 
the output conditional selection operators $\lambda_{g(x) \geq 0.0}$ and $\lambda_{g(x) < 0.33}$, and distance $d_{[0,0.33]}$, effectively measuring only the error of bins in the low category for which their mean labels are outside that category's interval.
We see that not only do all models have zero calibration error for the medium category, but some interventions have zero calibration error for all intervals.
This shows that all interventions produce outputs that better adhere to the specified Likert scale than the baseline.
Figure \ref{fig:IntervalCount} visualizes the number of instances in each interval.
All models output confidences in the medium interval less often than the two extremes, though the interventions produce more medium confidence outputs.
This indicates that the baseline model incurs interval-specific error that the interventions reduce by moving high-confidence outputs to the medium interval.

Together, Figures~\ref{fig:IntervalECE} and~\ref{fig:IntervalCount} show that 
some methods (TS, BCTS, and HB) model Likert categories especially well.
Methods that do incur category-specific error do so inconsistently (e.g. LS incurs higher mean-error in the ``high'' category than the ``low'', while FL and MU exhibit the opposite behavior).
Finally, this experiment shows that these interventions 
encourage more
 outputs in the ``medium'' confidence category
over
the baseline.
Such analysis can be beneficial when determining how likely the model is to output each category.

\textbf{Summary of Evaluation} In this section, we presented three case studies in classifier calibration, each motivated by use cases in which outputs of probabilistic classifiers are used to aid in making decisions.
We showed how to use the GECE framework to tailor calibration metrics to the contexts specified by each use case.
Beyond that, we found two common trends among the experiments.
First, improved performance over a baseline as measured Traditional ECE was not a strong indicator of improved performance in other variants of ECE used in our evaluation.
This motivates the use of context-specific ECE metrics when classifier usage is known a priori.
Second, techniques consistently reduced Traditional ECE by reducing the magnitude of the maximum classifier output, but in doing so increased the probability of other classes.
The techniques differed in which classes probability mass was increased, resulting in differing performance in ECE variants that consider more than the most probable output.
We believe this again motivates the use of more context-specific evaluation of calibration performance.

%


\section{Conclusion and Future Work}
In this work we propose a number of metrics to measure the context-specific calibration performance of probabilistic classifiers based on a more generalized statistical framing of expected calibration error.
Components that make up these metrics can be designed to measure error for isolated instances that can have application-specific importance, and to explicitly represent interpretations of classifier outputs that map to how the classifier will be used in practice. 
We evaluate a sampling of the state of the art in calibration, and show that our proposed metrics measure practical notions of calibration not captured by previously used metrics.


We have identified two noteworthy directions of future work.
First, one could extend our experiments to evaluate other kinds of calibration techniques
that do not neatly fit into the training-time or post-processing intervention categories
including
Monte Carlo Dropout~\citep{gal2016dropout}, Weight Uncertainty Neural Networks~\citep{blundell2015weight}, and Deep Ensembles~\citep{lakshminarayanan2017simple}.
Second, it is clear than none of the methods we evaluated were designed to calibrate their models according to the variety of metrics we proposed.
This motivates the need for techniques that are able to calibrate models according to 
more specific 
definitions of reliability.

\section{Acknowledgements}
Copyright 2022 Carnegie Mellon University.
This material is based upon work funded and supported by the Department of Defense under Contract No. FA8702-15-D-0002 with Carnegie Mellon University for the operation of the Software Engineering Institute, a federally funded research and development center.
The view, opinions, and/or findings contained in this material are those of the author(s) and should not be construed as an official Government position, policy, or decision, unless designated by other documentation.
This material is licensed under a  Creative Commons Attribution-Non-Commercial 4.0 International (CC BY-NC 4.0)
Carnegie Mellon® is registered in the U.S. Patent and Trademark Office by Carnegie Mellon University.


\bibliography{cal_eval}
\bibliographystyle{iclr2022_conference}


\clearpage
\appendix\label{appendix}
\section{Lenses and Distances}\label{sec:LensesNDist}
Here, we briefly discuss a lens and a distance that were omitted from the main paper. 

\subsection{Class Conditional}
To focus evaluation on a specific of interest $c$, one can use the \emph{class conditional} lens:
\begin{equation}
o(g(x)) = g(x)[c], t(\mathbf{y}) = \mathbf{y}\left[c\right] 
\label{eq:ClassCondLens}
\end{equation}
Here, both the classifier output and target label vectors are reduced to just their elements corresponding to class $c$.
This allows for the measurement of the calibration over a single class.
The authors of~\citet{nixon2019measuring} argued for the utility of this metric, and used it in their own evaluation.

\subsection{Class Weighted Error and Generalized Mahalanobis Distance}
In many applications, there is an inherent cost associated with particular classes.
For instance, in medical diagnosis, certain conditions are more severe than others and a classifier that erroneously predicts the presence of a severe condition can lead to unnecessary, invasive, and/or expensive treatments.
On the other hand, if a classifier fails to predict a severe condition when a patient does indeed have it, then the condition can go untreated.
For this reason, it can be useful to measure error with respect to cost associated with each class.
In \citet{Kumar2019VerifiedUC}, the authors propose to use the class conditional lens combined with the following weighted distance function to incorporate this idea into calibration error metrics:
\begin{equation}
    d(\mathbf{g},\mathbf{y}) = \left(\sum_{i}^{k}w_i\left(\mathbf{g}[i]-\mathbf{y}[i]\right)^2\right)^{1/2}
\label{eq:kumarloss}
\end{equation}
This distance function leads to an error that weighs the 
discrepancy between output and target 
for each class separately.
One can rewrite \eqref{eq:kumarloss} as the following:
\begin{equation}
    d(\mathbf{g},\mathbf{y}) = \left((\mathbf{g}-\mathbf{y})\mathbf{W}(\mathbf{g}-\mathbf{y})^T\right)^{1/2}
\label{eq:kumarloss_diag}
\end{equation}
Where, $\mathbf{W}$ is a $k \times k$ diagonal matrix, with $\left[w_1, w_2,...,w_k\right]$ down the diagonal.
More generally, $\mathbf{W}$ need not be diagonal.
Instead we can use a fully parameterized matrix $\mathbf{M}$, which makes \eqref{eq:kumarloss_diag} a generalized Mahalanobis distance:
\begin{equation}
    d(\mathbf{g},\mathbf{y}) = \left((\mathbf{g}-\mathbf{y})\mathbf{M}(\mathbf{g}-\mathbf{y})^T\right)^{1/2}
\label{eq:mahadist}
\end{equation}
In \eqref{eq:mahadist}, off diagonals represent weights corresponding to errors incurred between two classes (e.g. $\mathbf{W}_{i,j}$, weighs the error incurred when probability mass is allocated to class $i$ and the true class is $j$).
This can allow for calibration metrics that weigh particular kinds of confusion between classes.
For instance, one could assign lower weight the case when an instance of a malignant tumor has high classifier output for a ``benign tumor'' class than the case when it is not classified as a tumor at all.
The former may lead to a case where a tumor is extracted and found to be malignant by a pathologist, while the latter may result in the tumor not even being extracted.

Despite the potential utility, this metric formulation comes with a few caveats.
First, in order for \eqref{eq:mahadist} to be a proper distance metric, $\mathbf{M}$ must be positive semi-definite (PSD).
As such, it is not as simple as assigning intuitive weights to each element of $\mathbf{M}$ that corresponds to a particular kind of ``confusion'', as it might lead to a non-metric distance.
Second, the off diagonals do not have as intuitive of an interpretation as the diagonal entries do.
Even if the PSD constraint is not determined to be important when using the distance to model error, it is not obvious what exact values for the off-diagonals should be relative to each other.
One way to set these free parameters would involve
some level of exploratory analysis by developers using the metric in tuning the matrix with respect to examples where they would want error to be higher or lower than each other.
The alternative would be 
to learn $\mathbf{M}$ through expert feedback using some form of metric learning~\citep{bellet2015metric}. While we feel this is a promising metric formulation to be used for measuring calibration error, these complexities led us to leave a proper treatment of the topic to future work.

\section{Histogram Estimation}
\label{sec:HistEst}
In the course of our evaluation, we also analyzed the statistical properties of the histogram estimation involved in computing the expected value of the calibration error. Though often glossed over in many modern works in the deep learning literature that use the ECE as a calibration metric, we believe it is a key component to study in regards to computing any variant of the generalized ECE in an accurate and principled manner. Through our analysis we arrive at a practitioner heuristic for how to configure the simply parametrized algorithm used to perform the estimation, though we emphasize the need for continued study of this facet of the classifier calibration measurement problem.

\subsection{Binning Procedures}
A binning procedure determines the bins $\mathbb{B} = \left\{\mathcal{B}_1,...,\mathcal{B}_b\right\}$ for the histogram estimation involved in the error computation (stated in main work as \eqref{eq:ECEhist}, restated below, with a hat as $\widehat{ECE}$ \eqref{eq:ECEhist_hat}, to differentiate it from the true expected value statement of the $ECE$ \eqref{eq:ECE}).
The simplest choice is a \textit{uniform} scheme 
that partitions $\Delta^{k'-1}$ into evenly sized bins.
While uniform sized bins have intuitive individual interpretations because they slice the output space into fixed, identifiable regions, prior work in the statistical literature has shown that data-dependent binning schemes provide more accurate estimates~\citep{Nobel1996HistogramRE}.
As an alternative, \textit{adaptive} binning schemes have been proposed.\footnote{Adaptive schemes for ECE estimation have been employed in related work on classifier calibration such as~\citet{nixon2019measuring}, but only in binary settings.} 
Here, bin boundaries are determined by partitioning test instances $\mathcal{D}_{test} = \left\{\left(x_i,y_i\right)\right\}_{i=1}^{N}\mathtt{\sim}\ p\left(X,Y\right)$, such that bins contain (roughly) the same number of instances, but represent different sized subsets of $\Delta^{k'-1}$.
More specifically, a suitable algorithm that produces an $\gamma$-adaptive binning scheme will produce bins such that $\max_{i}{|\mathcal{B}_i}| \leq \gamma$.
Note that if $\gamma := \frac{N}{M}$, where $N = |\mathcal{D}_{test}|$, then the adaptive binning is data-dependent, and the fraction of instances in a bin is upper bounded by the choice of $M$, given the fixed number of a test samples of interest, $N$.
One appropriate algorithm for adaptive binning is to spatially partition instances using a $k$-$d$ tree.
We adopt an implementation originally proposed for this specific purpose in~\citet{vaicenavicius2019evaluating}.

\subsection{A Heuristic to Ensure Estimator \emph{Consistency}}

Despite the fact that ECE metrics computed using binned estimation schemes
can exhibit undesirable statistical consistency 
~\citep{Kumar2019VerifiedUC}, the framework remains popular due to its direct interpretability with respect to calibration conditions. 
Encountering this shortcoming in our experiments, we found that controlling the bias of the histogram estimator is key in achieving accurate ECE estimates in practice.

\begin{figure}
  \begin{minipage}[c]{0.49\linewidth}
    \includegraphics[width=\linewidth]{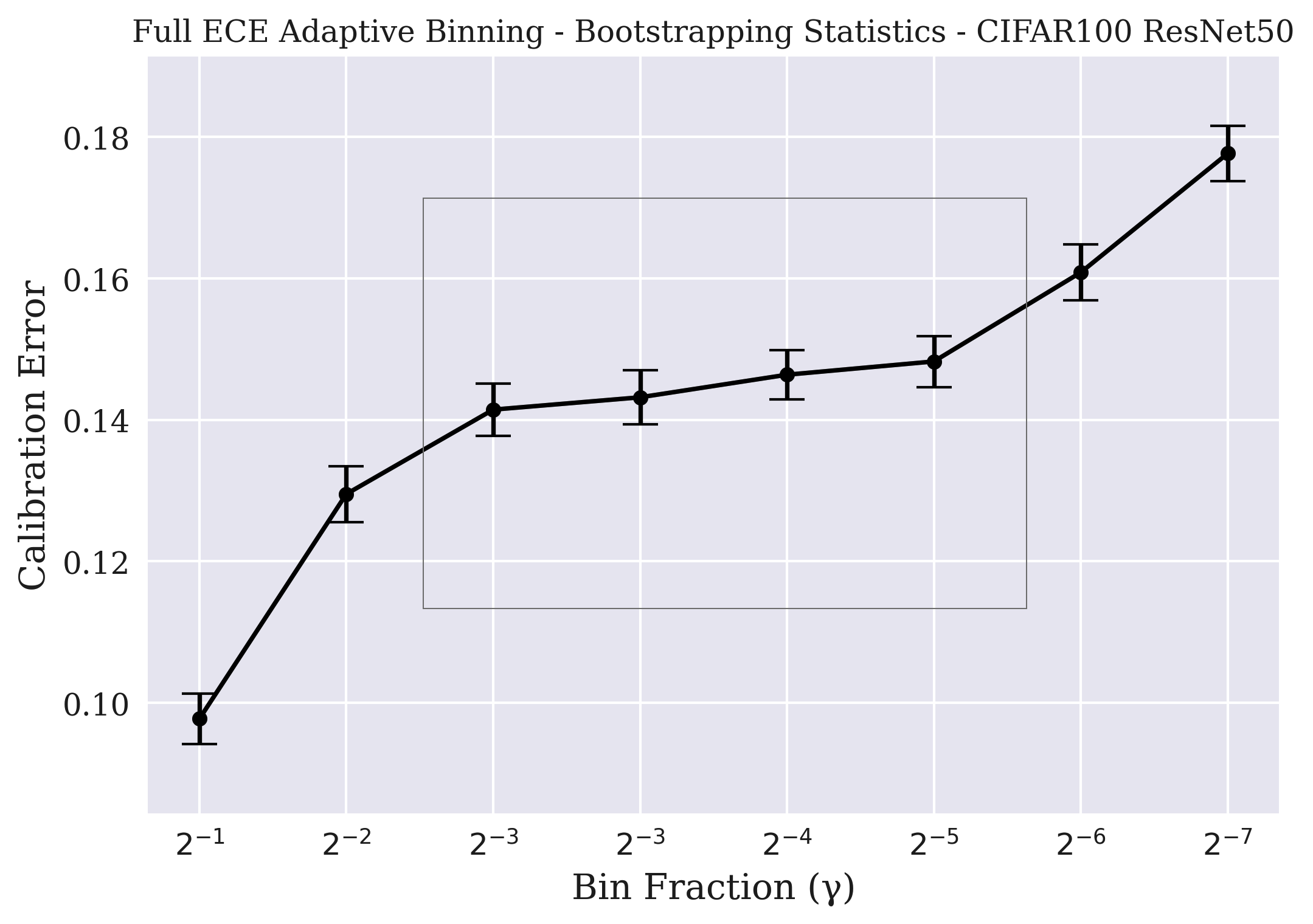}
    \caption{Full ECE versus adaptive bin fraction parameter $\gamma$}
  \label{fig:biases-adaptive}
  \end{minipage}
  \hfill
  \begin{minipage}[c]{0.49\linewidth}
  \includegraphics[width=\linewidth]{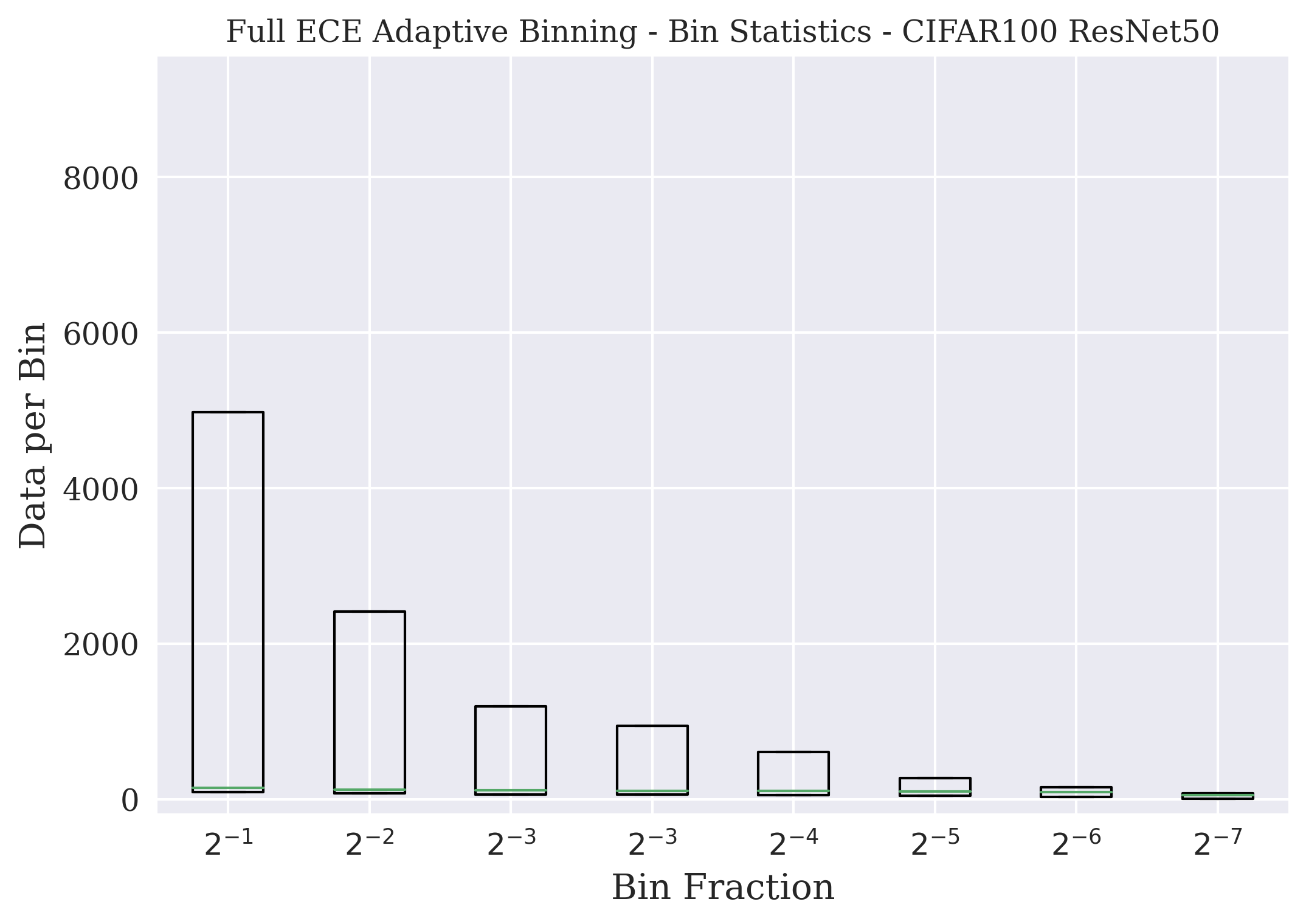}
    \caption{Median, min, and max, points per bin in the set of bins generated by the adaptive scheme versus bin fraction parameter $\gamma$}
  \label{fig:bin-stats-adaptive}
  \end{minipage}
\end{figure}

Based on our evaluation, and the foundations established in prior work~\citep{vaicenavicius2019evaluating}, we recommend setting the adaptive binning parameter $\gamma$ to balance biases in the estimator.
Figure \ref{fig:biases-adaptive} plots the full ECE of a classifier versus $\gamma$, over $1000$ bootstrapped re-samples of test data.
Note that in contrast to all other error bars in this work, which are based on the result of multiple training trials, the error bars on Figure~\ref{fig:biases-adaptive}, are constructed using this bootstrapped resampling with replacement (N draws from N points). The bars displayed show the standard deviation of the metric value over those $1000$ resamples, at each $\gamma$ setting.
We observe that ECE ``stabilizes'' around $\gamma = 2^{-4}$, effectively balancing the negative bias arising from a binning that is too coarse grained, and the positive bias from lacking sufficient samples per bin.
We use this heuristic to determine settings for $\gamma$ in our experiments.

\subsection{Statistical Properties of the Histogram Estimator}
In this section we elaborate on a number of characteristic behaviors exhibited by generalized ECE metrics that are computed using binned, plug-in estimation schemes. In particular, we formalize bias and variance trends using a combination of arguments from two previous works in the calibration literature as well as basic intuition about density estimation. Here we provide a consolidated discussion of asymptotics and implications posed in those works as well as empirical evidence of their relevance in practice. The goal here is to motivate the importance of studying the estimation component of the framework due to its direct effect on the quality of your metric values. Especially when utilizing reliability conditions that are more expressive than the traditional Top-1 ECE, the strength of the statements that can be made about the effectiveness of a given model calibration technique directly depends on the quality of your error estimations. 

For reference in the sections to follow, the formal definitions for the Expected Calibration Error and the Histogram Estimation of the ECE are (re)stated below as Equations (\ref{eq:ECE}) and (\ref{eq:ECEhist_hat}), respectively. Though the notation is slightly overloaded, in the main work, and the appendices, if not explicitly stated, the ``calibration error'', or ``ECE'', when referred to as an empirically measurable quantity, is by definition, the \textit{estimated} error (\ref{eq:ECEhist_hat}).


\begin{equation}
ECE(g) := \mathbb{E}_{X,Y\mathtt{\sim}P\left(X,Y\right)}d\left(\mathbb{P}\left(Y | g\left(X\right)\right),g\left(X\right)\right)
\label{eq:ECE}
\end{equation}

\begin{equation}\label{eq:ECEhist_hat}
    \widehat{ECE}(g) = \sum_{\mathcal{B}\in\mathbb{B}} \frac{|\mathcal{B}|}{|\mathbb{B}|}d\left(\bar{\mathbf{g}}_{\mathcal{B}},\bar{\mathbf{y}}_{\mathcal{B}}\right)
\end{equation}

\subsubsection{Convergence}

When using the generalized ECE framework, it is suggested that given ``sensible'' choices for binning procedures used to generate $\mathbb{B} = \left\{\mathcal{B}_b\right\}_{b=1}^{M}$, ones that result in smaller and smaller bins as the size of the finite sample $N = |\mathcal{D}_{test}|$ grows, the estimate of the calibration error (\ref{eq:ECEhist_hat}), converges
to the true calibration error (\ref{eq:ECE})~\citep{Nobel1996HistogramRE}. A formalization of this convergence assumption in the sufficiently many samples and finely-grained binning scheme limit is provided as Theorem 1 in~\citet{vaicenavicius2019evaluating}, restated with some simplification below.

Given continuity of the distance function and smoothness of the ``perfect re-calibration'' mapping of $\mathbf{g} \to \mathbf{y}$ implied by the data and learned model (i.e. the posthoc transformation applied to the model that would be required to make it perfectly output the label distribution everywhere), with $diam(\mathcal{B}_b) := \sup_{x_i,x_j \in \mathcal{B}_b}\lVert x_i - x_j \rVert_{2}$, if $\lim_{M \to \infty}{\max_{b}{(diam(\mathcal{B}_b)})} = 0$, we have:

\begin{equation}\label{eq:convergence}
\lim_{M,N \to \infty}{\widehat{ECE}(g)} = ECE(g)
\end{equation}

Both uniform and data-dependent schemes can satisfy the required conditions of for (\ref{eq:convergence}), however, appropriate adaptive data-dependent schemes are expected to provide better estimator consistency for a fixed sample of a given size. Thus, for most sample sets, adaptive schemes are expected to produce more accurate error estimates. As mentioned previously, Figure~\ref{fig:biases-adaptive} shows the ECE as a function of the adaptive binning granularity parameter $\gamma$, and additionally Figure~\ref{fig:bin-stats-adaptive} gives an summary of the spread of points-per-bin membership counts for the actual binnings realized by the algorithm at each $\gamma$ setting. Recall that $\gamma = \frac{N}{M}$, and as such, the ratio in those figures varies from large to small, left to right along the x-axis, implemented via increasing the number of bins, $M$, which we refer to as the direction of increasing \emph{binning granularity}.



\subsubsection{Negative Asymptotic Bias as $N \to \infty$}
In addition to the general dependence of the estimator quality on the type of binning procedure discussed in the previous section, another important relationship between the true ECE and the empirical realization computed via histogram estimation is the underestimation, or negative asymptotic bias, of the binned estimator. Proposition 3.3 in~\citet{Kumar2019VerifiedUC} formalizes this for binary lenses and Theorem 2 in~\citet{vaicenavicius2019evaluating} does the same for arbitrary lenses, conditioned on a finite, fixed data-\emph{independent} (such as uniform, or similar) partitioning and basic continuity of the distance function:

\begin{equation}
\lim_{N \to \infty}{\widehat{ECE}(g)} \le ECE(g)
\end{equation}

The intuition for the source of the negative bias incurred by any fixed binning scheme as samples increase, is that the mean calculation in each bin absorbs the overconfidence and underconfidence incurred in different subsets of that bin. This results in a loss of accumulated error in each individual bin, and thereby a lower overall error value. This bias is a general consequence of the discretization to the mean, and is thereby expected independent of the specific binning procedure or the distance function chosen for your ECE metric. The left extreme of  Figure~\ref{fig:biases-adaptive} demonstrates the empirical presence of this bias mode when the binning granularity is \emph{too coarse}.

\subsubsection{Positive Asymptotic Bias as $M \to \infty$}
There exists a second, competing, asymptotic bias that arises, especially when the distance function used is based on the $\ell_1$ vector norm (such as TVD). In constructing the $\widehat{ECE}$, we compute the bin-wise mean outputs $\bar{\mathbf{g}}_{\mathcal{B}}$ and mean targets $\bar{\mathbf{y}}_{\mathcal{B}}$ to approximate the local, or per bin, values that are ``plugged-in'' to the distance function of the calibration error. As a result this type of estimator is sometimes referred to as a \textit{plug-in estimator}.

The combination of the $\ell_1$ norm and the plug-in procedure results in a positive bias when the binning scheme is too fine-grained for the number of samples available. For a fixed sample of size $N$: 

\begin{equation}
\lim_{M \to \infty}{\widehat{ECE}(g)} \ge ECE(g)
\end{equation}

The argument in Appendix G of~\citet{Kumar2019VerifiedUC} is that the positive bias of the binned calibration error arises due to the nature of the absolute value. If you consider any fixed, finite, sample set, the difference between the two plug-in estimates $\bar{\mathbf{g}}_{\mathcal{B}}$ and $\bar{\mathbf{y}}_{\mathcal{B}}$ computed from the samples will be non-zero even for a perfectly calibrated model where, in expectation, the per bin mean output and mean target are actually equivalent (zero error according to (\ref{eq:ECE}), the non-estimated $ECE$). Thus, for the estimated $\widehat{ECE}$, these overestimations accumulate in the overall error value in the limit. The right extreme on Figure~\ref{fig:biases-adaptive} demonstrates the empirical presence of this bias mode when the binning granularity is \emph{too fine}. 


\subsubsection{Sample Complexity of the Variance}

\begin{wrapfigure}{r}{0.5\textwidth}
  \centering
    \includegraphics[width=\linewidth,trim={0 0.2cm 0 0},clip]{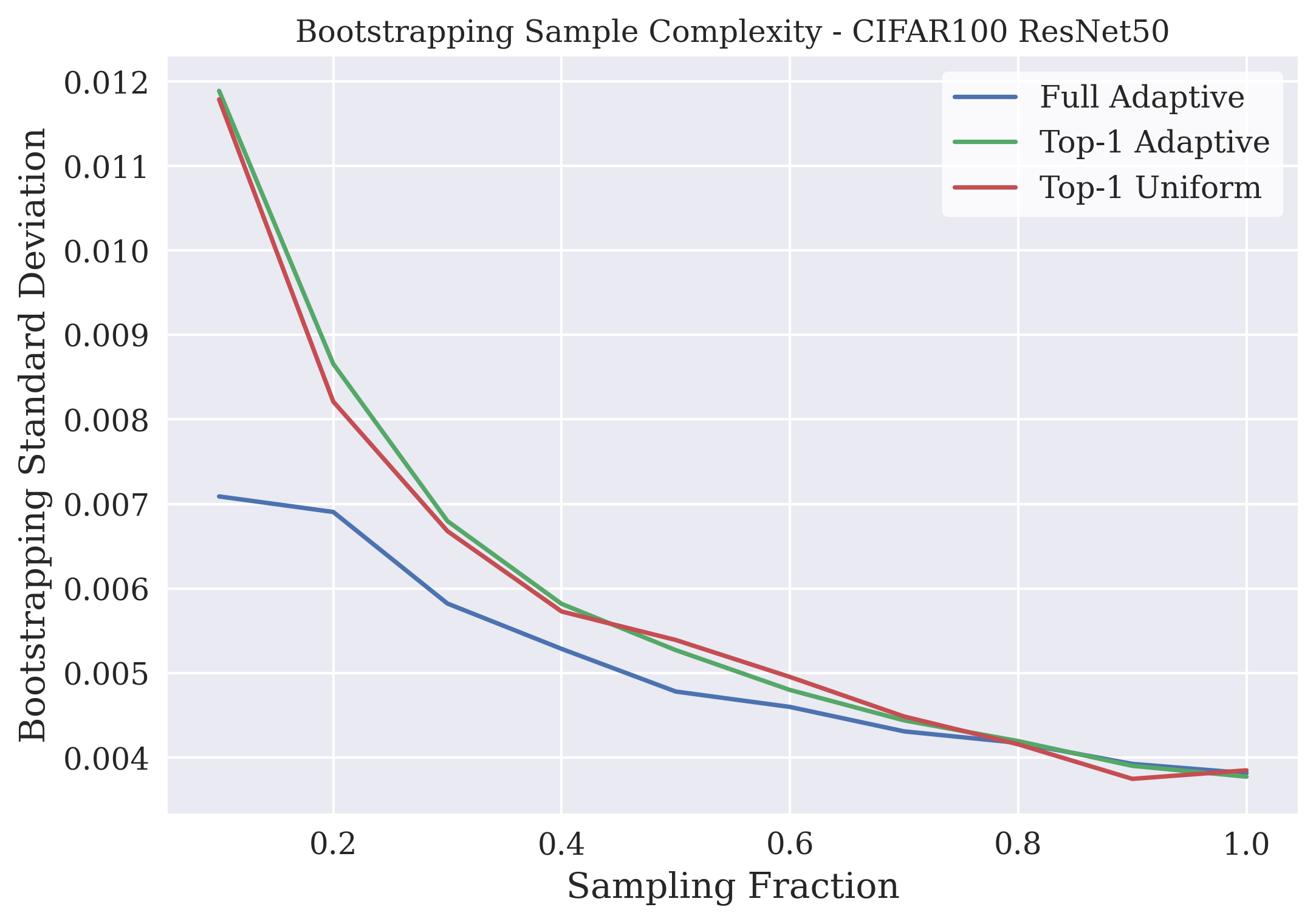}
    \vspace{-0.7cm}
    \caption{The standard deviation of the generalized ECE on 1000 bootstrap resamples with replacement at a sample fraction of $\frac{n}{N}$}
  \label{fig:variance-sample-complexity}
  \vspace{-0.8cm}
\end{wrapfigure}

Additionally, following a basic statistical intuition for the bais-variance tradeoff expected when performing any histogram density estimation, we hypothesize that in the limit of a binning scheme that is too fine-grained for the number of samples available, there should be an increase in variance of the $\widehat{ECE}$. We empirically verify this by fixing the adaptive binning parameter $\gamma$, and modifying the bootstrap resampling process used to create error bars in Figure~\ref{fig:biases-adaptive} such that instead of drawing $N$ times from the $N$ points, we draw $n$ times, creating the axis series for $\frac{n}{N}  \in [0.1, 0.2, ..., 1.0]$. On the low end of the range, this artificially puts the estimator into the extremely low sample availability regime for the given model and data manifold, and produces increased variance of $\widehat{ECE}$ and this is visualized in Figure~\ref{fig:variance-sample-complexity}. However, we note that the absolute standard deviation is still relatively small ($<0.01$ where the absolute error is $\sim 0.14$, with spread of $0.08$ over the $\gamma$ range). 
We conclude that this hypothesized behavior, increased variance due to low sample availability, is not the estimator consistency facet that a practitioner should worry about when working with data sets with standard values for $N$, and $C$ the number of classes, reasonable class balance, and performant model architectures (CIFAR-10/100/ImageNet and deep convolutional networks). However, we do note that narrower lenses on the model output, like Top-1, seem to exhibit higher sensitivity (increase spread of the variance over $\gamma$ range) in the low sample-availiability regime than the full ECE, suggesting this could be more of an issue in practice due to the ubiquity of the Top-1 variant.

\subsection{Practitioner Prescription and Future Work}

The adaptive binning granularity parameter setting, $\gamma = 0.1$, is suggested 
as a baseline for a $k$-$d$ tree based algorithm on standard data sets.
This is a practitioner guideline derived by interpreting the empirical results across our generalized ECE types and selection of data sets and models in the context of the theoretical foundations from previous work. By sweeping the granularity parameter for your binning algorithm and identifying a region of relative stabilty in the metric value (the boxed region of Figure~\ref{fig:biases-adaptive}), a practitioner has some evidence that their metric of interest is not dominated by either asymptotic behavior mode - the negative bias arising from a binning that is too coarse, or the positive bias from lacking sufficient samples per bin when computing plug-in means. That being said, based on our results, we posit the existence of tighter, or at least more fully characterized, bounds between the histogram estimated $\widehat{ECE}$ and true $ECE$ when working with standard data contexts and corresponding values for $N$ and $M$. More comprehensive analyses of data dependent binning algorithms, their parametrization, and their sample complexity, especially for the more expressive and understudied ``full'' ECE lens, is an important direction of further research.

\section{Additional Experimental Details}\label{sec:AddExpDets}

In order to provide for technical transparency, reproducibility of our results, and further experimentation based on our work, we describe our experimental setup in further detail below. However, we direct readers to our source code release (upon publication) as enumeration of the full details is better suited to the model and algorithmic definitions provided by our implementation, as well as the parameter settings listed in human-readable configuration files.

\subsection{Data Sets and Models}
The three neural network architectures used in the evaluation are a ResNet50~\citep{he2016cvpr}, a MobileNetV2~\citep{sandler2018mobilenetv2}, and a DenseNet100 and DenseNet201~\citep{huang2017densely}. The implementations of each are ported from PyTorch Vision (\url{https://github.com/pytorch/vision}) with modifications to the DenseNet architecture based on the Lua Torch repository originally released with the DenseNet paper (\url{https://github.com/liuzhuang13/DenseNet}). The data sets used across each of the different experimental settings are roughly divided into the ``standard'' data sets for training models on the multi-class classification task from scratch 1) CIFAR-100~\citep{krizhevsky2009learning}, and ImageNet~\citep{deng2009imagenet}, and 2) those which we take a model first pretrained on ImageNet, and then finetune it on a second classification task - CBIS-DDSM~\citep{lee2017curated, shen2019deep}, ETHEC, Crisis-MMD, and COVID-CXR-2~\citep{pavlova2021covid}.

For data sets that have a published \textit{train}, \textit{test} and \textit{validation} split, the validation split is used, else a fraction of the training set is withheld as the validation set. For each model and data set pairing, any architectural changes required to reconcile model input and output dimensionality are made, and then a series of data pre-processing and augmentation transformations are applied to the data, and a loss function, network optimizer, and training schedule are specified.

We first establish \textit{base} models, where no calibration intervention is applied, and then create corresponding versions where each intervention is applied individually. We configure model components and set hyperparameters using a combination of the details from each model's introduction paper and the papers in which each individual calibration technique is proposed. We also survey the benchmark accuracy numbers found in a results meta-repository, \url{paperswithcode.com}, to make sure that our configuration for each model architecture achieves sufficient accuracy (relative to similar architectures) on each data set in the evaluation that is outside of the original model paper's set of published results. The extensive enumeration of all training configuration details and hyperparameters is ill suited for prose or even tabular format, but for each (model, data set, TT intervention) triplet, of which there are $3 \times 6 \times (3+1)$ including the base model, we provide in our code release, a human readble YAML configuration file that describes all parameters required to replicate each of the accuracy, generalized ECE, and other metric values we report. Architectural changes to models for handling different image sizes and class counts can be inspected in the architectures section of the library in our source code release.

\subsection{Interventions/Techniques}

Train-time interventions are implemented as changes to the training configuration, and therefore require changes to one or more of the base model's training parameters. Post-processing interventions are implemented as an additional module, post-processor $pp$, taking as input the logits or softmax layer outputs of the base model. The Train-time interventions introduce an extra characterizing hyperparameter each, controlling the \emph{strength} of the technique, that must be set for each model and data set pairing it's applied to. The PP intervention modules require a set of learned parameter(s) to be optimized with respect to the negative log likelihood of the composition $pp \circ g$ on the the held-out validation set, $nll(pp(\mathbf{g}),\mathbf{y})$. We set each Train-time intervention's strength hyperparameter to the value that minimizes traditional, or top-1 ECE without sacrificing more than $5\%$ accuracy over the base model by performing a grid search over a fixed range of values chosen based on the settings reported in each technique's original paper, as well as subsequent works that utilize the technique.

Train-time interventions and their respective strength hyperparameter:
\begin{itemize}
    \item Label smoothing - smoothing $\alpha$
    \item Mix-up - mix-up $\alpha$
    \item Focal loss - $\gamma$ schedule
\end{itemize}
Post-processing interventions and their learned parameter(s):
\begin{itemize}
    \item Temperature scaling - temperature $T$
    \item Bias-corrected temperature scaling - $T,\mathbf{b}$
    \item Histogram binning - number of bins
\end{itemize}




We scope our evaluation to a sampling of techniques in the Train-time and Post-processing intervention classes, but note that the analysis done in this work can be extended to other techniques. We publicly release our evaluation code to facilitate the evaluation of techniques not included in this work.





\section{Additional Experimental Results}
\label{sec:AddExpRes}

\subsection{Top-1 Accuracy}
In Tables 1-7 we report all the accuracy values for each baseline model as well as the models with each intervention for all experiments.
All values reported are the means over five trials, the exception being ImageNet, whose accuracy values are from 1 or 2 trials, depending on the architecture.

\subsection{Results Supplementing Those from Main Paper}
In this section, we show some additional results that supplement the experiments shown in the main body of the paper.
Each subsection here extends the subsection of the same name in Section~\ref{sec:empiricalEval} of the main paper.
\subsubsection{Top-$k$ Lens}
\begin{figure}
    \centering
    \includegraphics[width=0.8\linewidth]{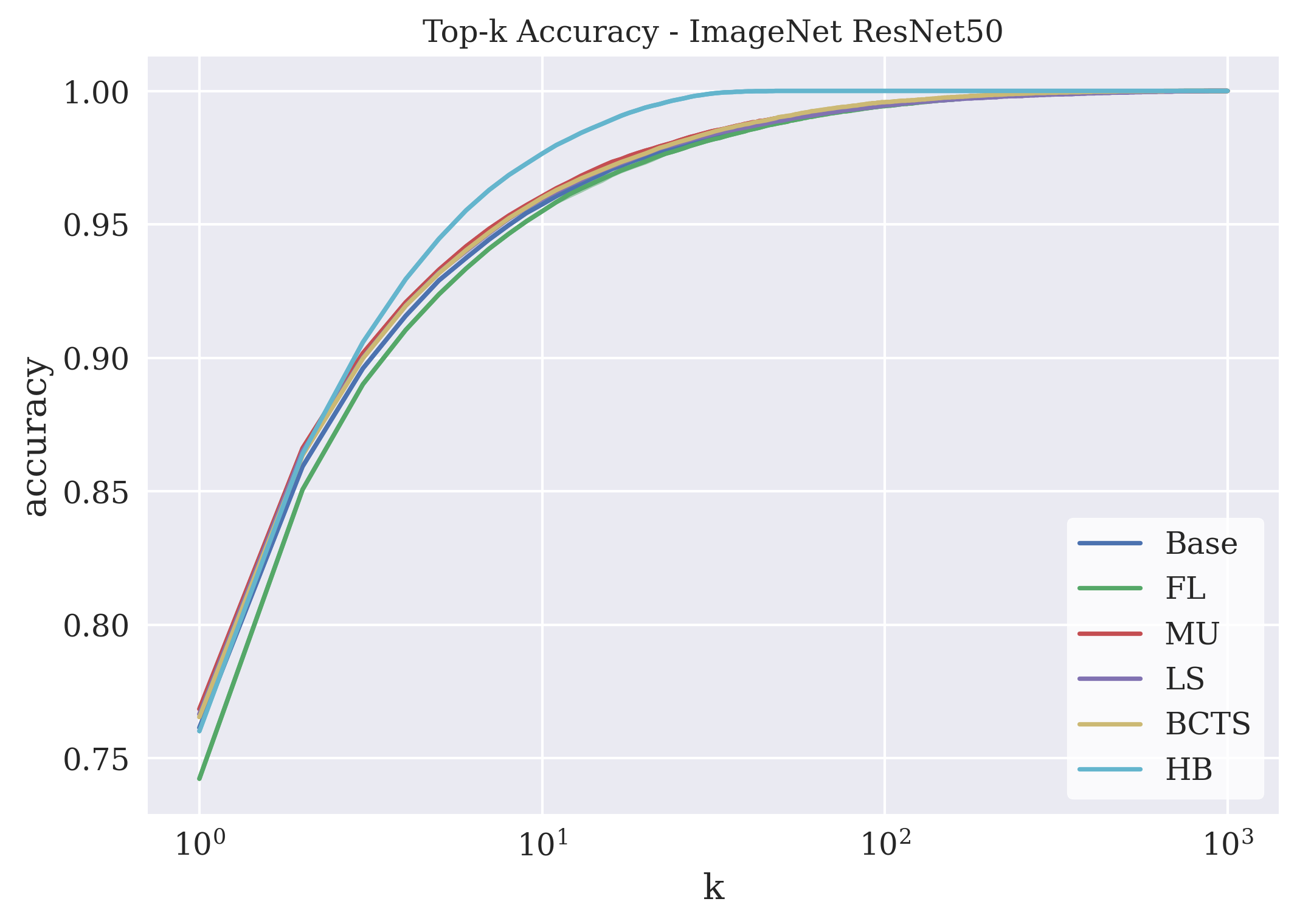}
    \caption{Top-$k$ test set accuracy versus $k$ (All Interventions, ResNet50, ImageNet)}
    \label{fig:TopKAcc}
\end{figure}

\begin{figure}
    \centering
    \includegraphics[width=0.8\linewidth]{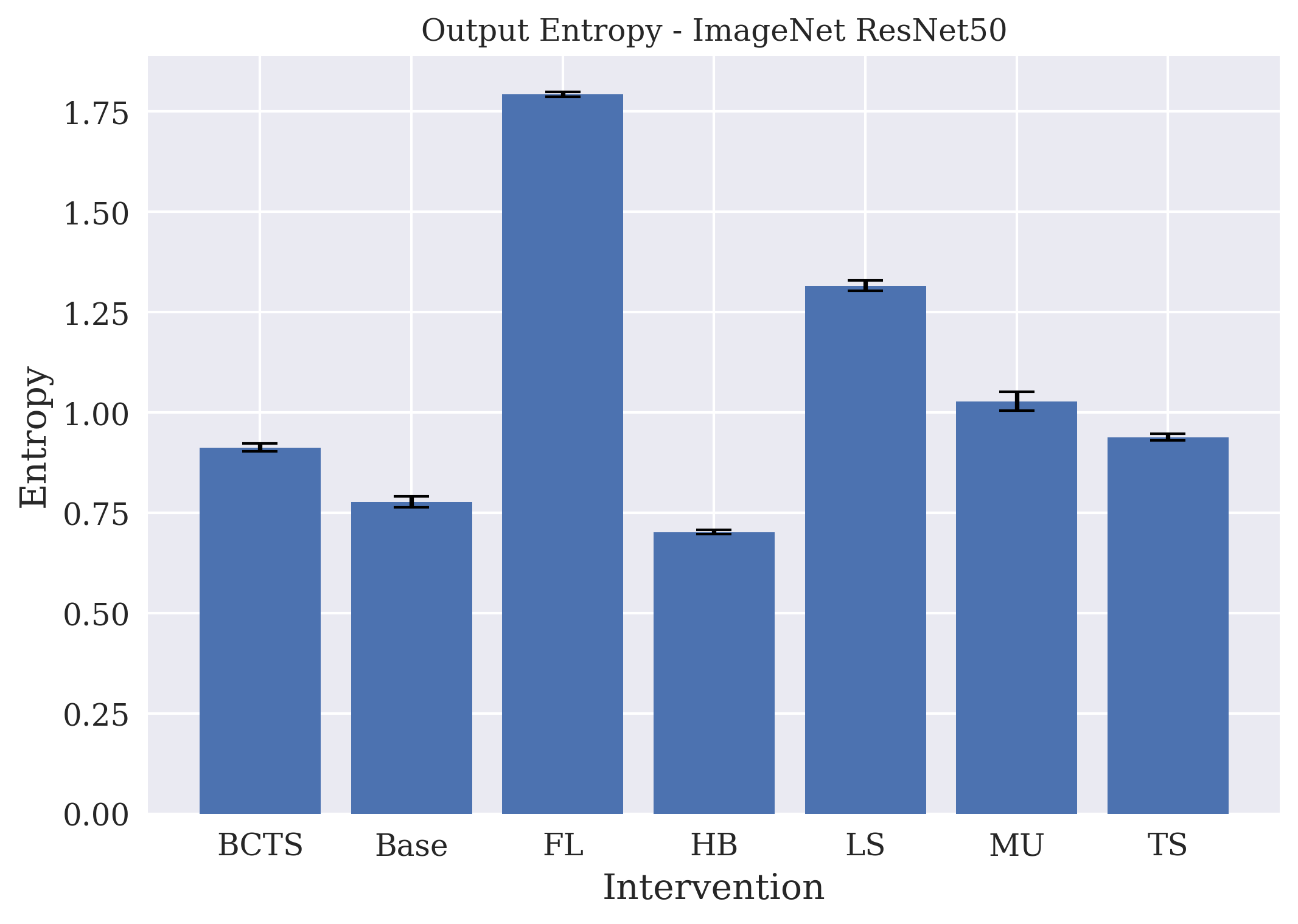}
    \caption{Mean entropy of outputs on test set (All Interventions, ResNet50, ImageNet)}
    \label{fig:TopKEnt}
\end{figure}

Figure \ref{fig:TopKAcc} plots the top-$k$ accuracy of a ResNet50 model composed with each calibration intervention as a function of $k$ on the ImageNet test set.
Here, we see that for the base model the correct class is almost always in the top 100 highest output classes.
Because of this, any technique that allocates probability mass to classes $>100$ likely incur error for top-$k$ ECE metrics for $k>100$.
This can explain why some methods that more evenly distribute probability mass incur error at higher $k$ values.
Figure \ref{fig:TopKEnt} shows the mean entropy of of the output distributions of all interventions applied to ResNet50 on the ImageNet test set.
We see that all techniques (with exception of HB) increase the entropy, showing that they effectively push their output distributions towards uniform, as reported for some of the interventions in prior work.

\subsubsection{Grouping Lens and Label Selection}
\begin{figure*}
    \centering
    \includegraphics[width=\linewidth]{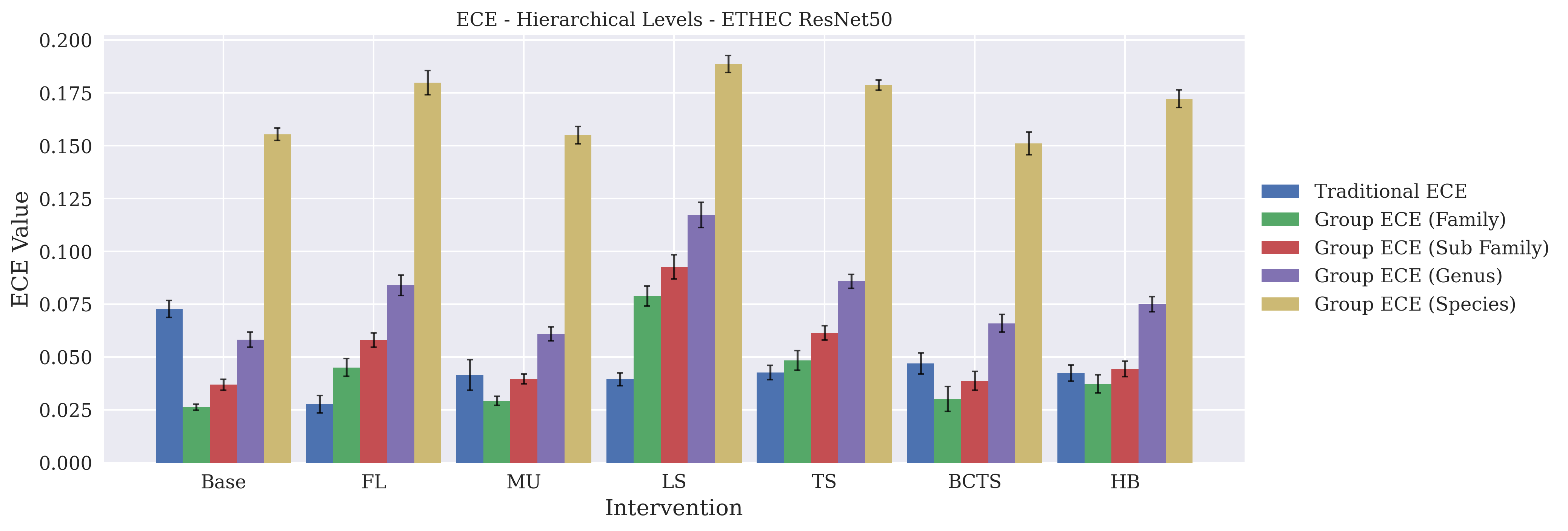}
    \caption{Group-wise ECE, various groupings (All Interventions, ResNet50, ETHEC)}
    \label{fig:MultiGroup}
\end{figure*}

Figure \ref{fig:MultiGroup} shows the Group-wise ECE (no selection operators) of all interventions with groupings at all levels of the hierarchy from ``family'' to ``species''.
Here, we see that as the groupings get finer in granularity, the ECE increases, starting ``family'' ECE (same as the green ``family'' bars in Figure~\ref{fig:GroupingECE} of the main paper), and ending with the ``species'' grouping, which is equivalent to a full ECE measure (the classes used to train the models were based on the species labels).
We see that as the groupings get finer in granularity, the ECE increases, which is a natural effect of inducing more specific classification problems.

\subsubsection{Inter-Interval Distance and Output Selection} 
\begin{figure*}
    \centering
    \includegraphics[width=\linewidth]{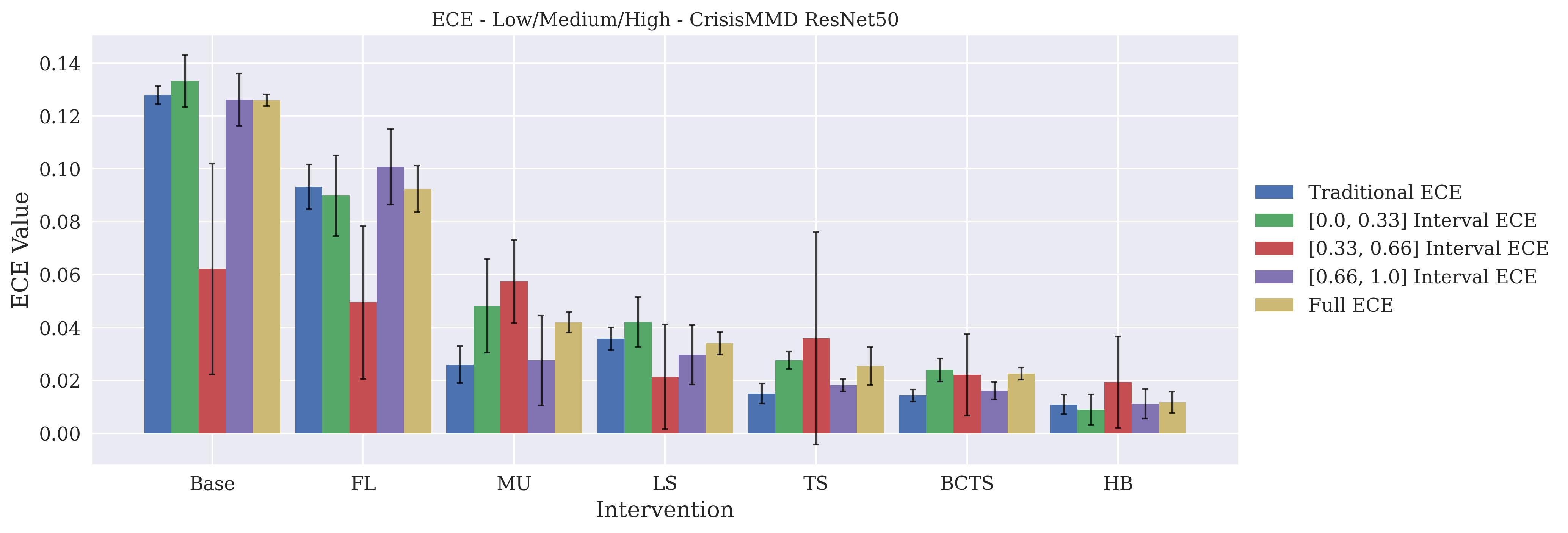}
    \caption{Interval-wise ECEs, TVD Distance (All Interventions, ResNet50, CrisisMMD)}
    \label{fig:IntervalTVD}
\end{figure*}

Figure \ref{fig:IntervalTVD} shows the same results as Figure \ref{fig:IntervalECE} of the main paper, except instead of using inter-interval distance to measure inter-interval error, we use standard total variation distance (TVD) to measure both inter- and intra- interval error.
Stated another way, using TVD instead of inter-interval distance here removes the effect of suppressing error that occurs only within a bin's interval.
The main result of note is that all methods encounter significant error in the [0.33, 0.66] ``medium'' interval when measured by TVD, while they incur no error when the inter-interval distance is used. 
This highlights that bins in the medium interval for all methods have no mean labels that fall outside of the medium interval, but they are not exact estimates of their mean outputs.

\subsection{Results with Additional Network Architectures}
\begin{figure*}
    \centering
    \includegraphics[width=\linewidth]{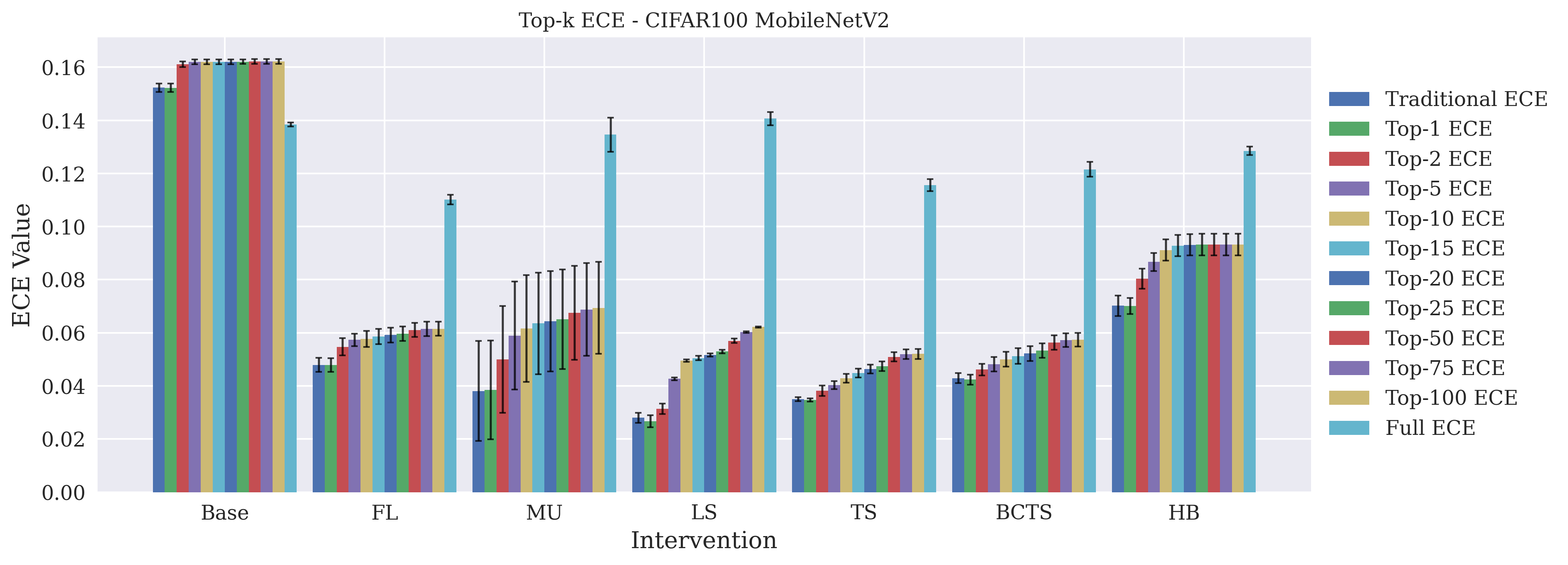}
    \caption{Top-$k$ ECE (All Interventions, MobileNetv2, CIFAR-100)}
    \label{fig:TopKECEMobile}
\end{figure*}

\begin{figure*}
    \centering
    \includegraphics[width=\linewidth]{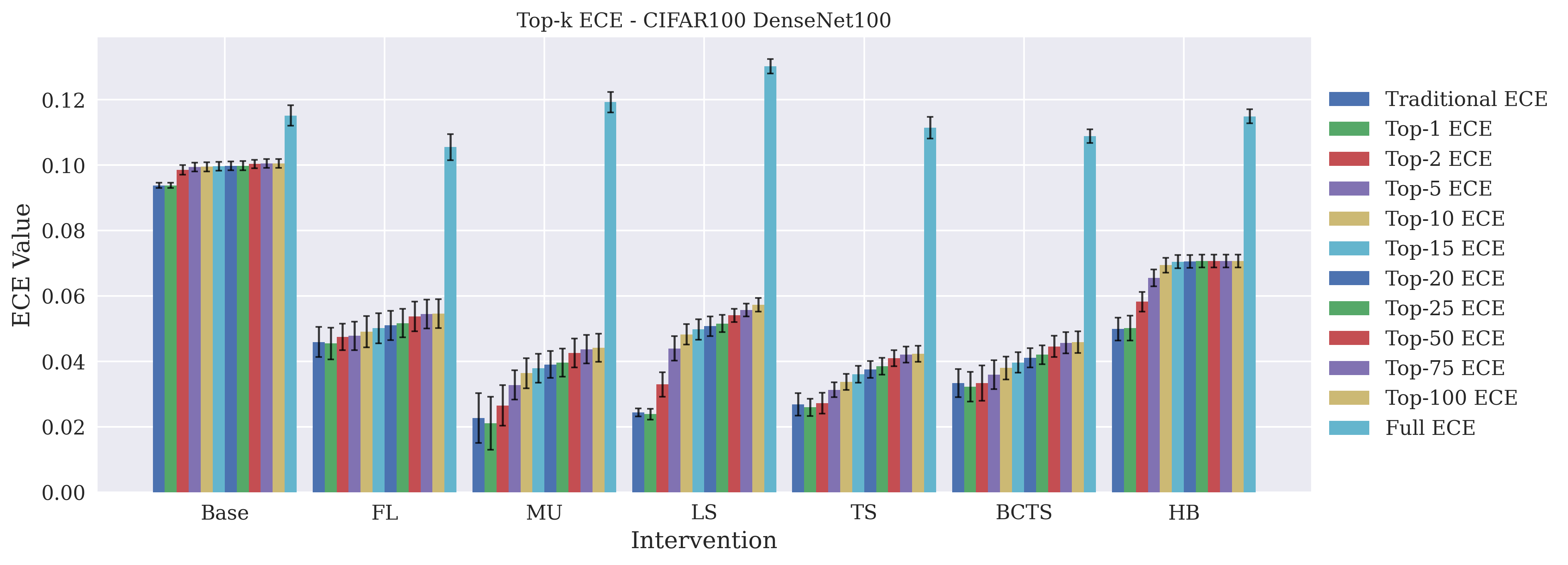}
    \caption{Top-$k$ ECE (All Interventions, DenseNet100, CIFAR-100)}
    \label{fig:TopKECEDense}
\end{figure*}

\begin{figure}
    \centering
    \includegraphics[width=0.65\linewidth]{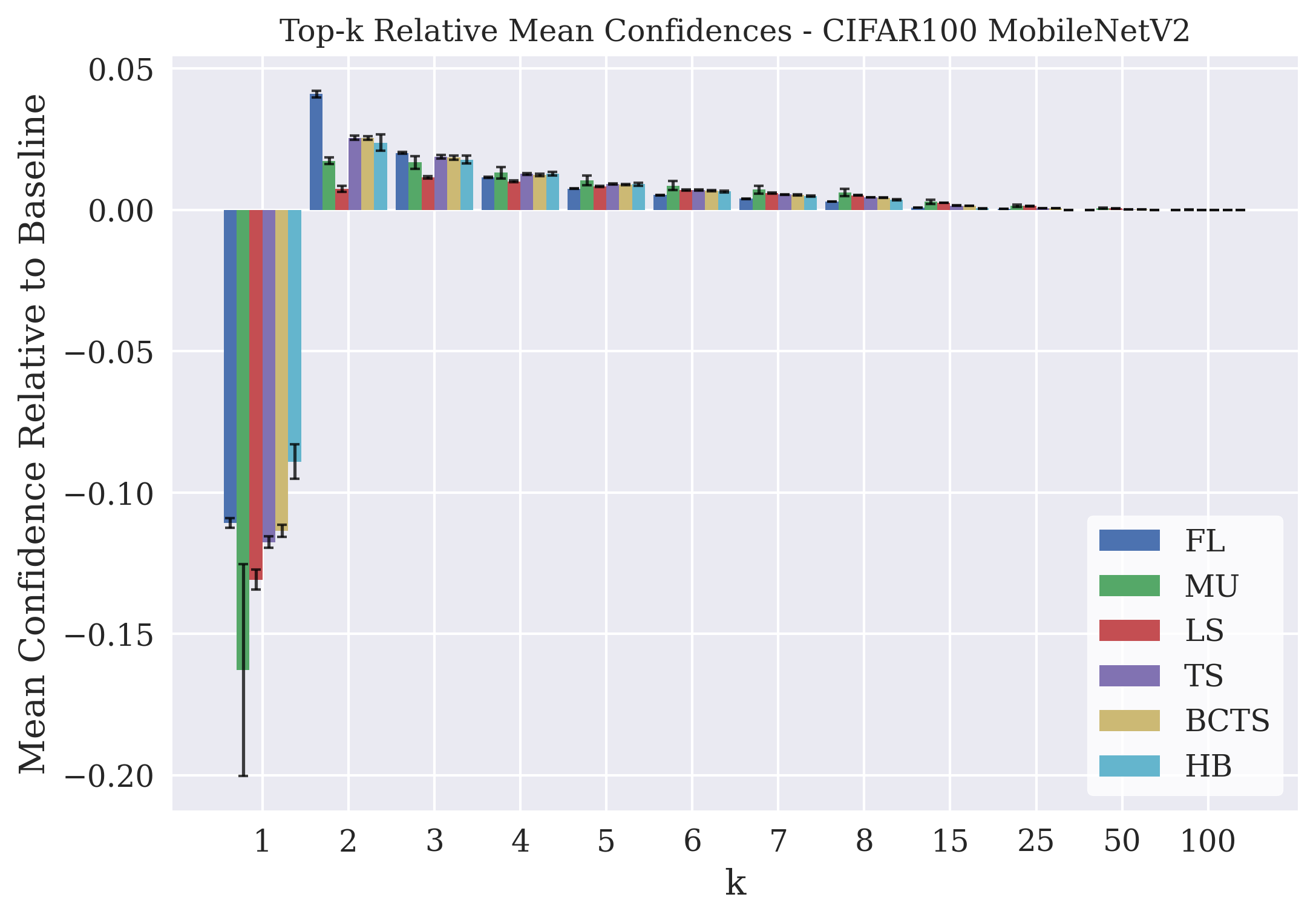}
    \caption{Mean Top-$k$ confidences, relative to baseline (All Interventions, MobileNetv2, CIFAR-100)}
    \label{fig:TopKConfMobile}
\end{figure}

\begin{figure}
    \centering
    \includegraphics[width=0.65\linewidth]{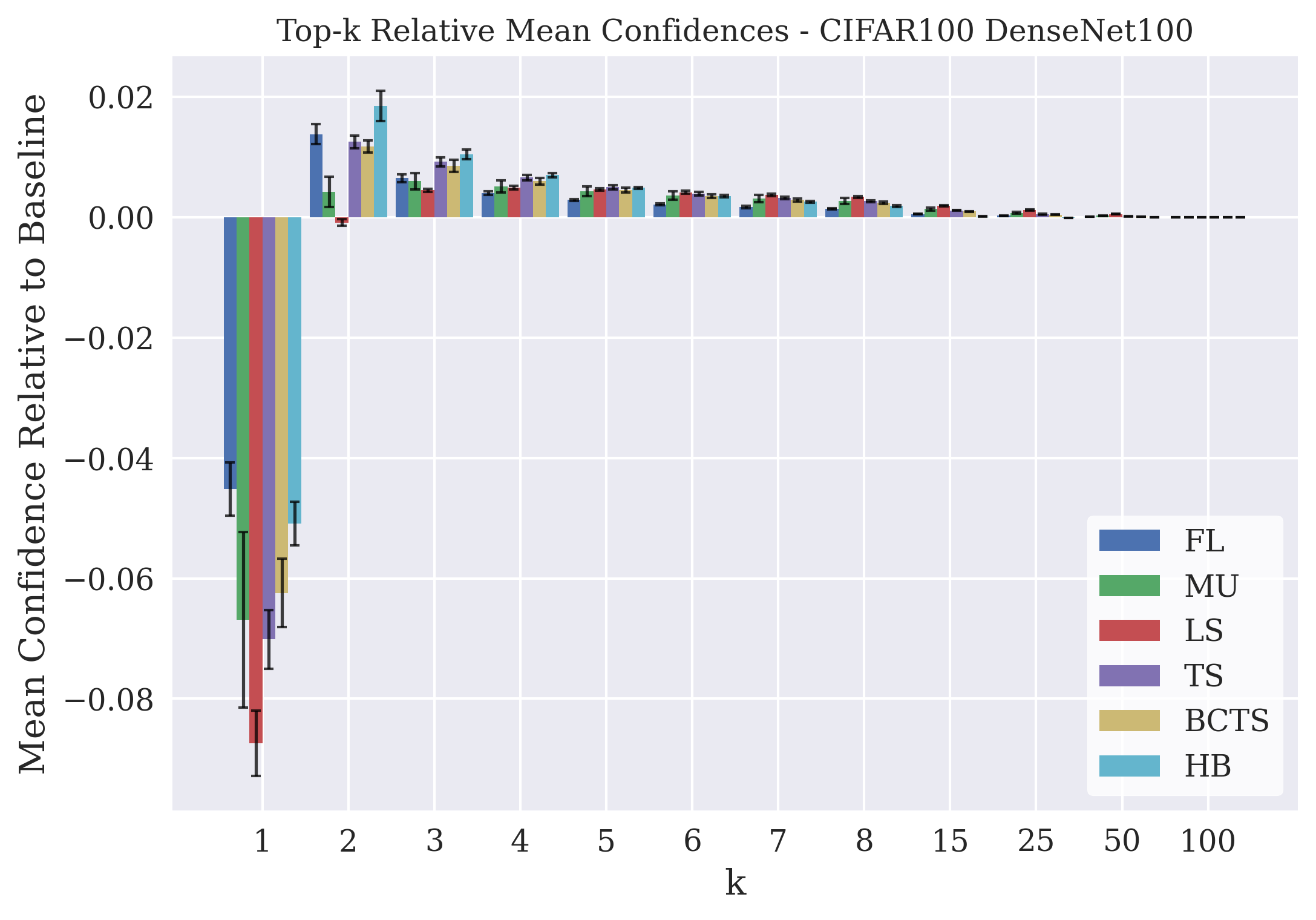}
    \caption{Mean Top-$k$ confidences, relative to baseline (All Interventions, DenseNet100, CIFAR-100)}
    \label{fig:TopKConfDense}
\end{figure}

\begin{figure*}
    \centering
    \includegraphics[width=\linewidth]{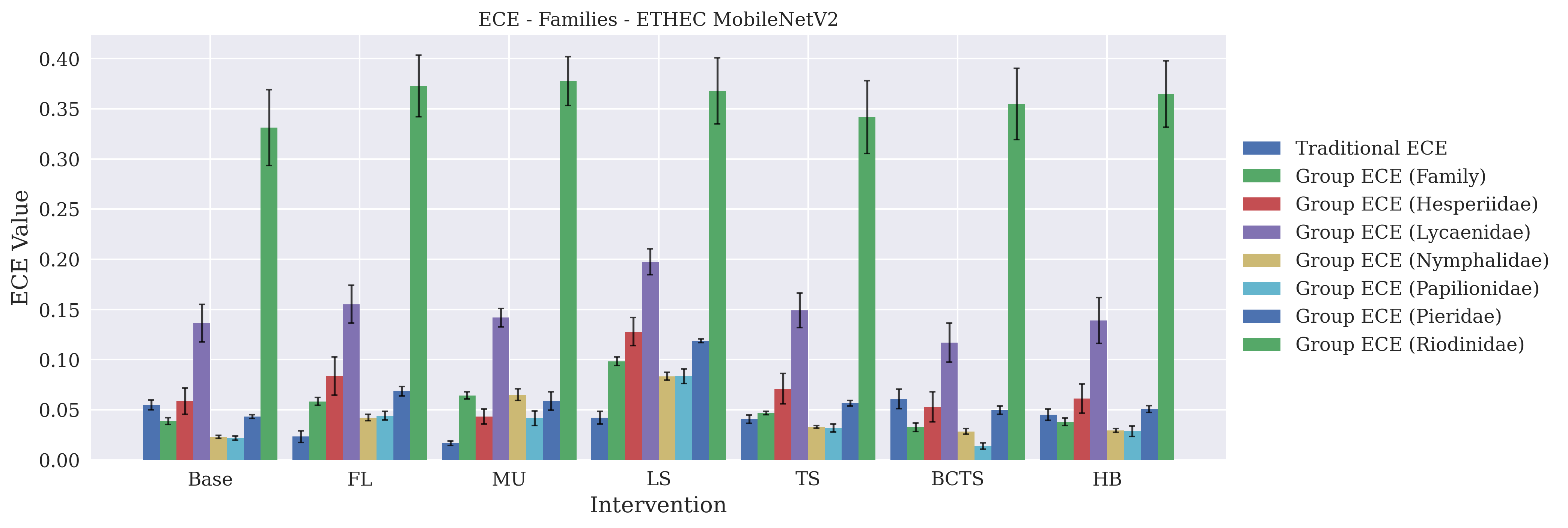}
    \caption{Group-wise ECEs, (All Interventions, MobileNetv2, ETHEC)}
    \label{fig:GroupingECEMobile}
\end{figure*}

\begin{figure*}
    \centering
    \includegraphics[width=\linewidth]{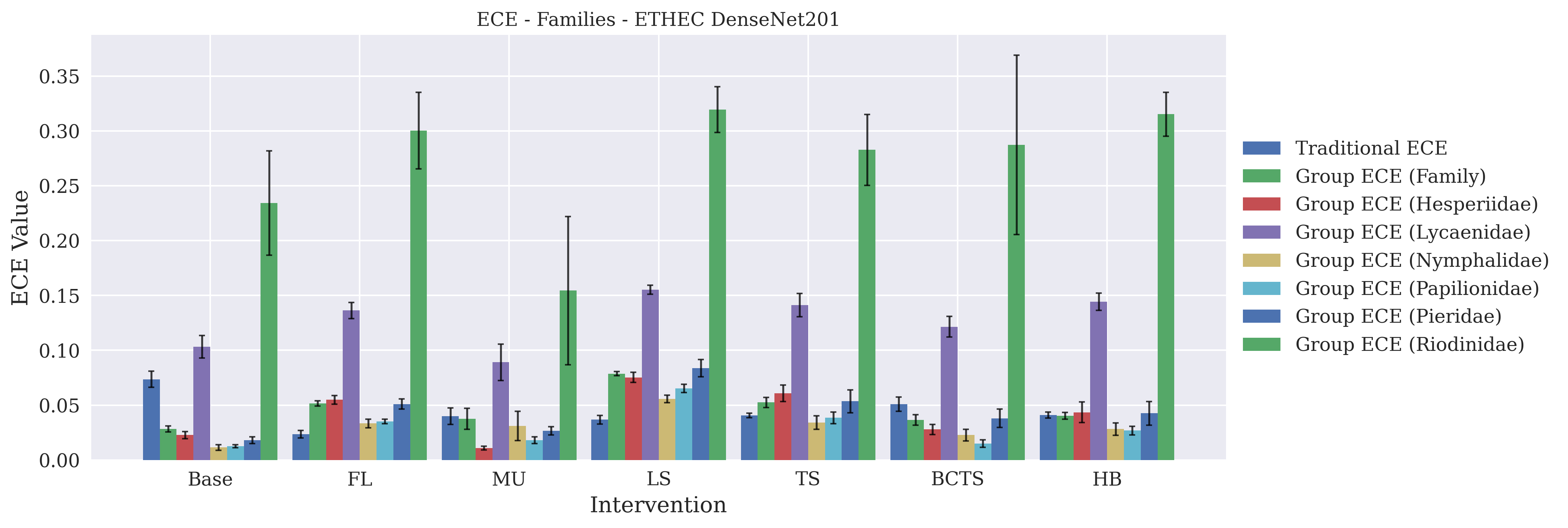}
    \caption{Group-wise ECEs, (All Interventions, DenseNet201, ETHEC)}
    \label{fig:GroupingECEDense}
\end{figure*}

\begin{figure}
    \centering
    \includegraphics[width=0.8\linewidth]{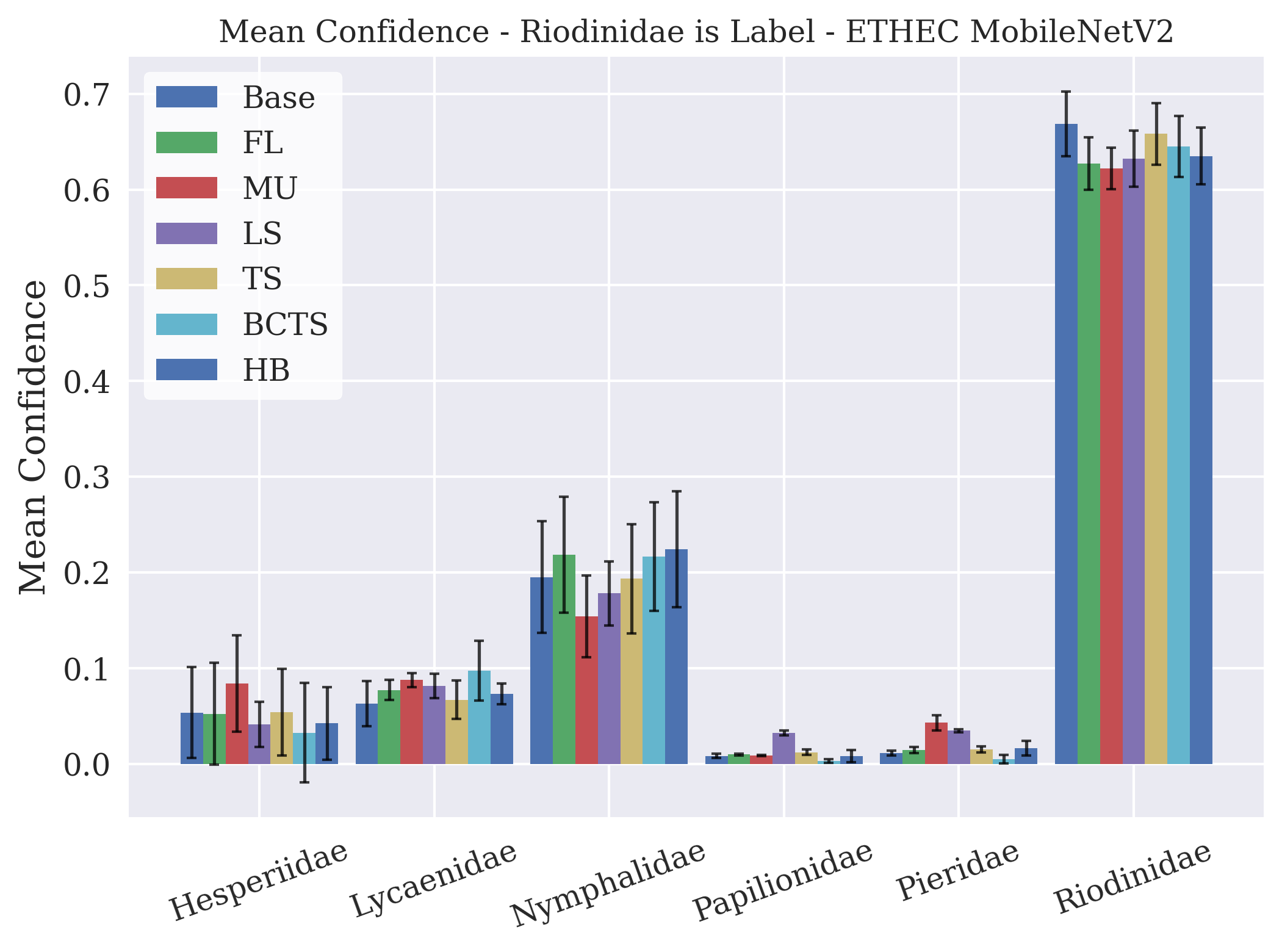}
    \caption{Mean family-wise confidence, conditioned on ``Rhiodinidae'' group (All Interventions, MobileNetv2, ETHEC)}
    \label{fig:GroupingConfMobile}
\end{figure}

\begin{figure}
    \centering
    \includegraphics[width=0.8\linewidth]{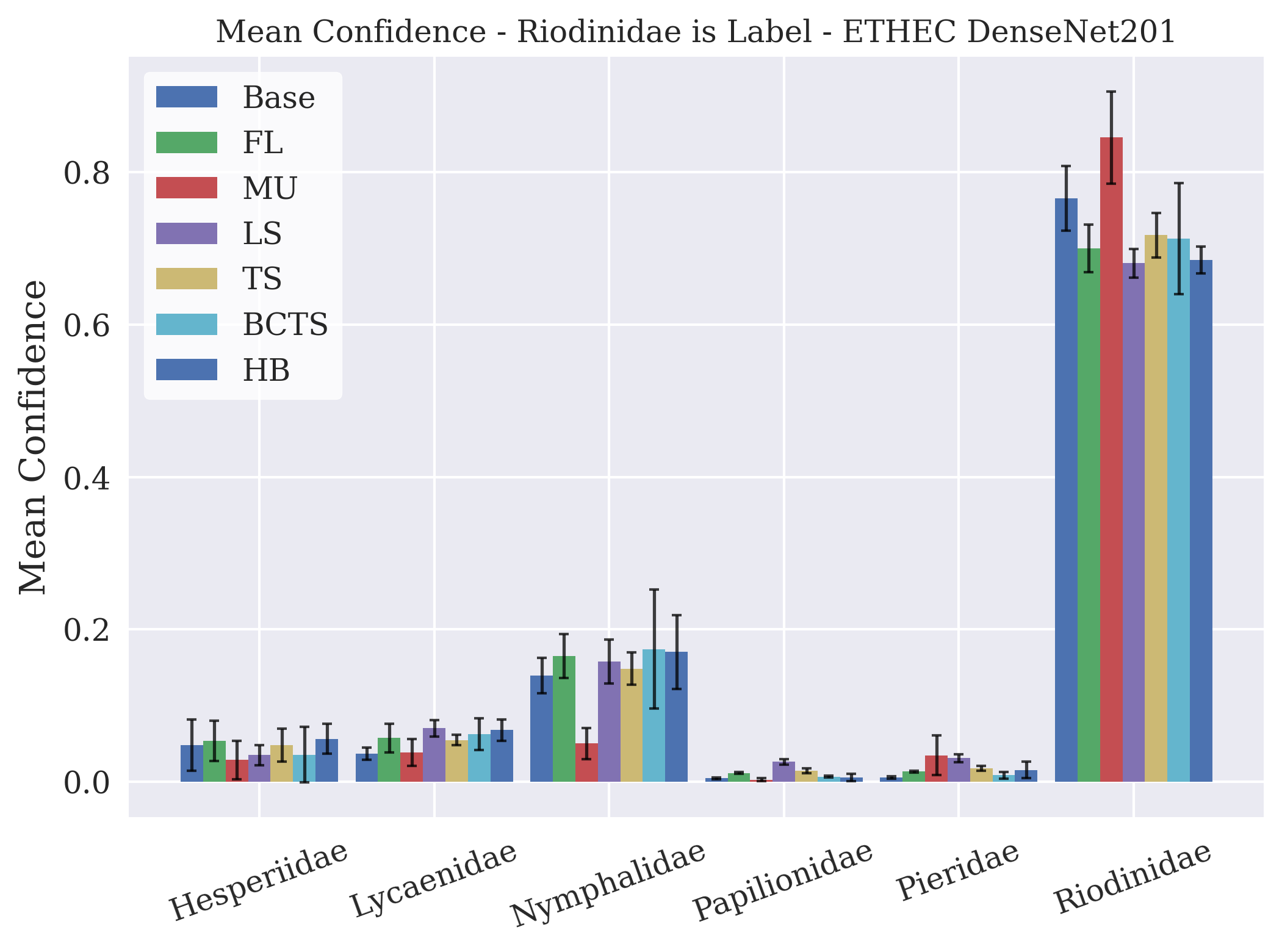}
    \caption{Mean family-wise confidence, conditioned on ``Rhiodinidae'' group (All Interventions, DenseNet201, ETHEC)}
    \label{fig:GroupingConfDense}
\end{figure}

\begin{figure*}
    \centering
    \includegraphics[width=\linewidth]{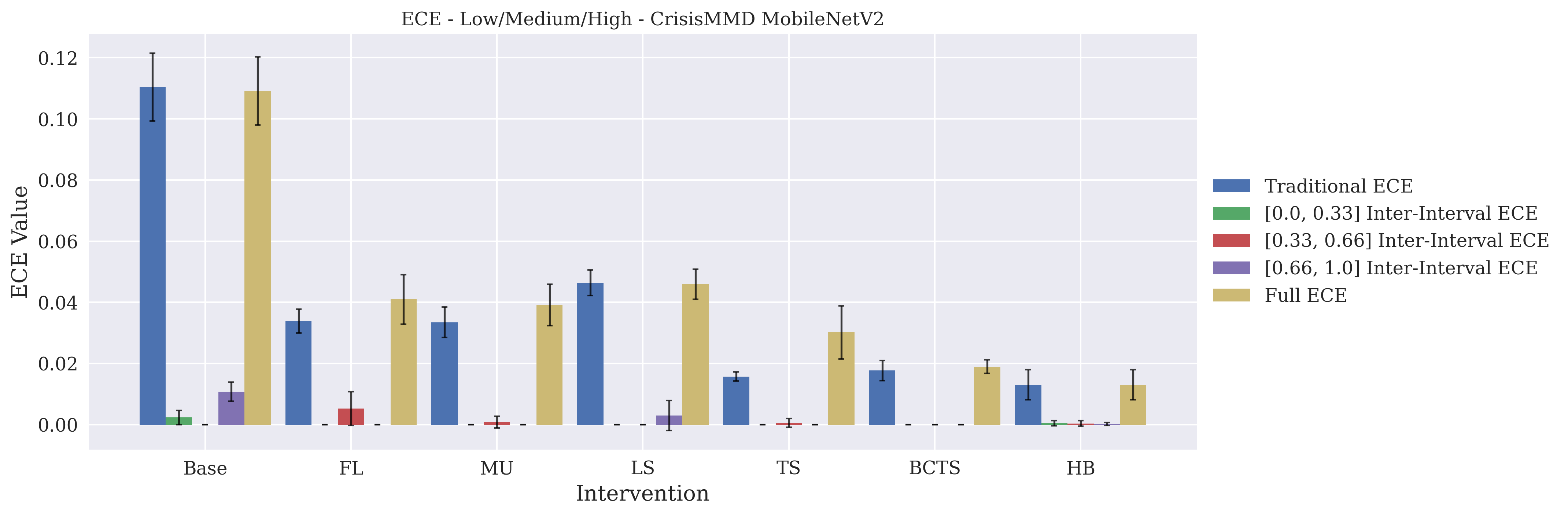}
    \caption{Interval-wise ECEs (All Interventions, MobileNetv2, CrisisMMD)}
    \label{fig:IntervalECEMobile}
\end{figure*}

\begin{figure*}
    \centering
    \includegraphics[width=\linewidth]{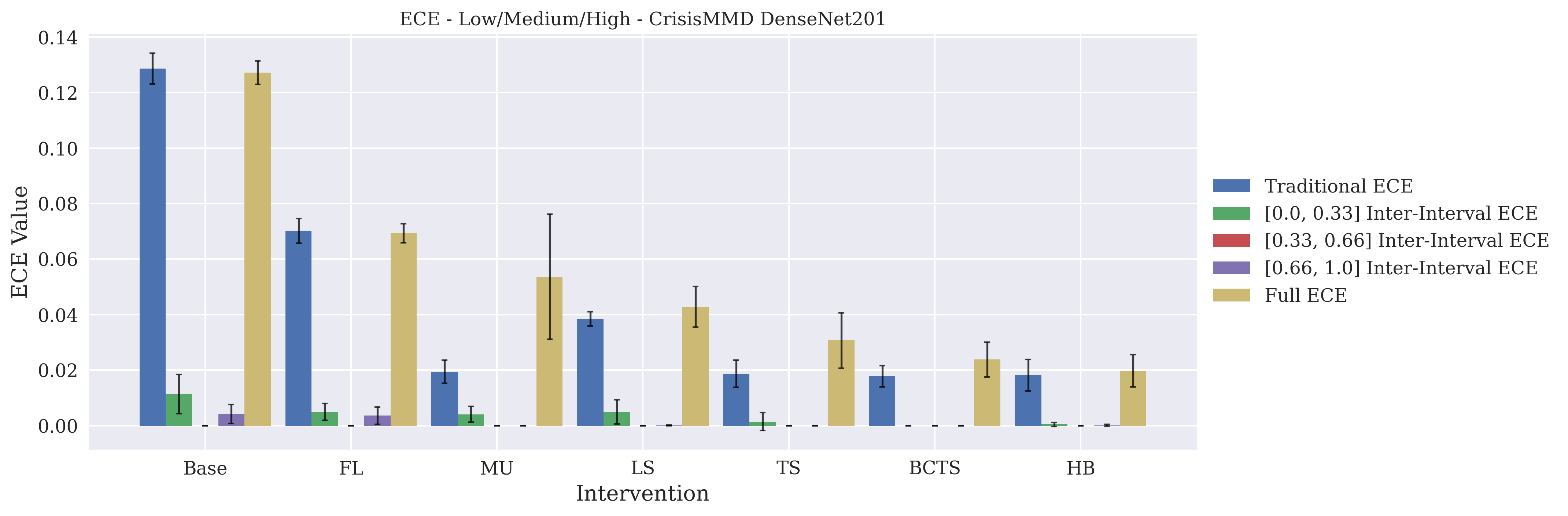}
    \caption{Interval-wise ECEs (All Interventions, DenseNet201, CrisisMMD)}
    \label{fig:IntervalECEDense}
\end{figure*}

\begin{figure}
  \centering
    \includegraphics[width=0.8\linewidth]{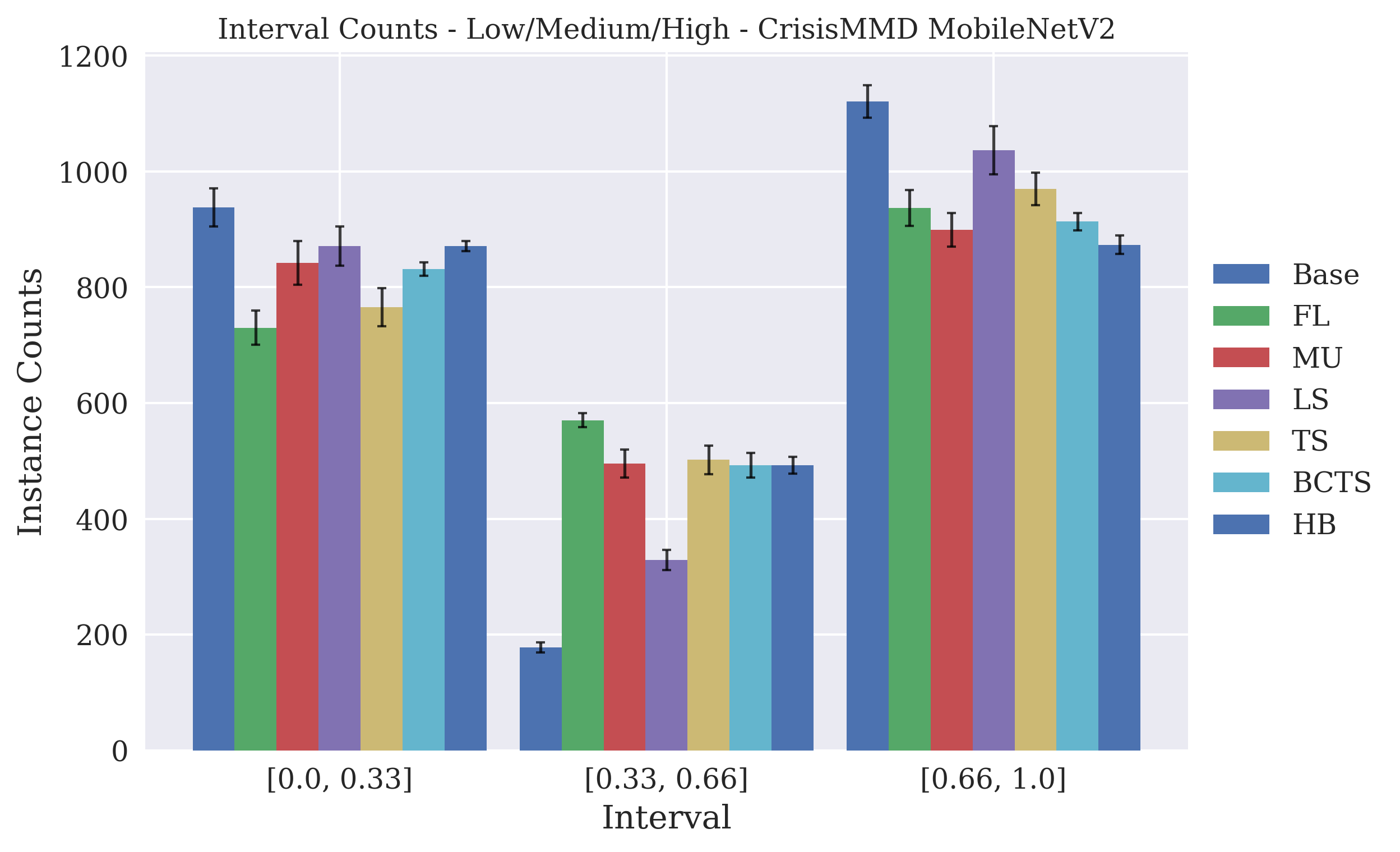}
    \caption{Test instances per interval (all interventions, MobileNetv2, CrisisMMD)}
  \label{fig:IntervalCountMobile}
\end{figure}

\begin{figure}
  \centering
    \includegraphics[width=0.8\linewidth]{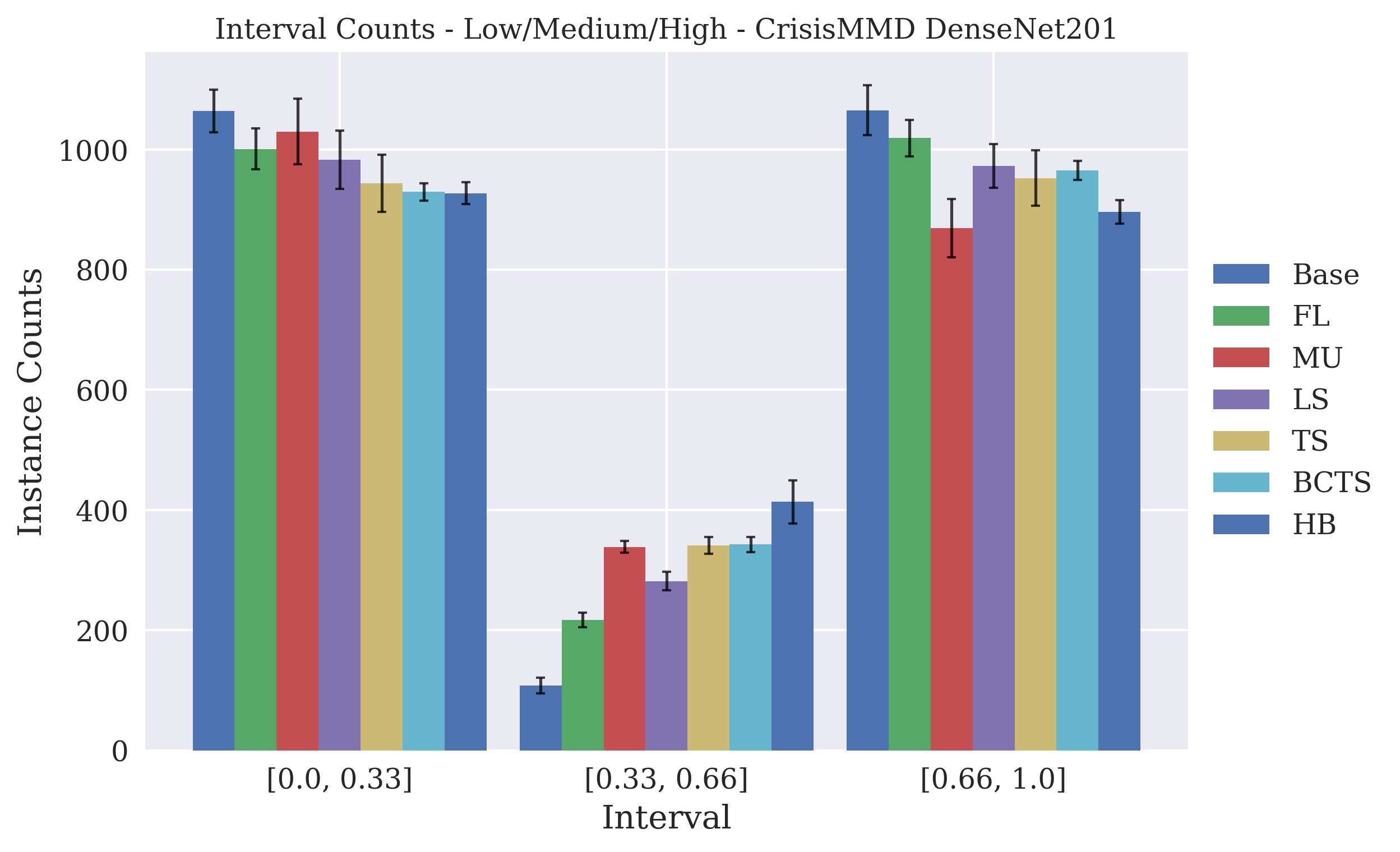}
    \caption{Test instances per interval (all interventions, DenseNet100, CrisisMMD)}
  \label{fig:IntervalCountDense}
\end{figure}

Figures \ref{fig:TopKECEMobile}-\ref{fig:IntervalCountDense} show the same experiments used to generate the figures in Section~\ref{sec:empiricalEval} in the main paper with the only difference being the neural network architecture used.
Here, we use MobileNetv2 and a 100 layer DenseNet (DenseNet100) instead of the 50 layer residual network (ResNet50) used in the main paper.
We did this for the sake of rigor, and found that the results reported in the main paper for ResNet50 also largely hold for these two architectures.
We believe this is evidence that our results generalize across a variety of popular neural network architectures.

\subsection{Results using Additional Data Sets}
For each of the three experiments we provided in the main paper, we ran a similar experiment on an additional data set.
In this section we show these additional results.

\subsubsection{Top-$k$ Lens}

\begin{figure*}
    \centering
    \includegraphics[width=\linewidth]{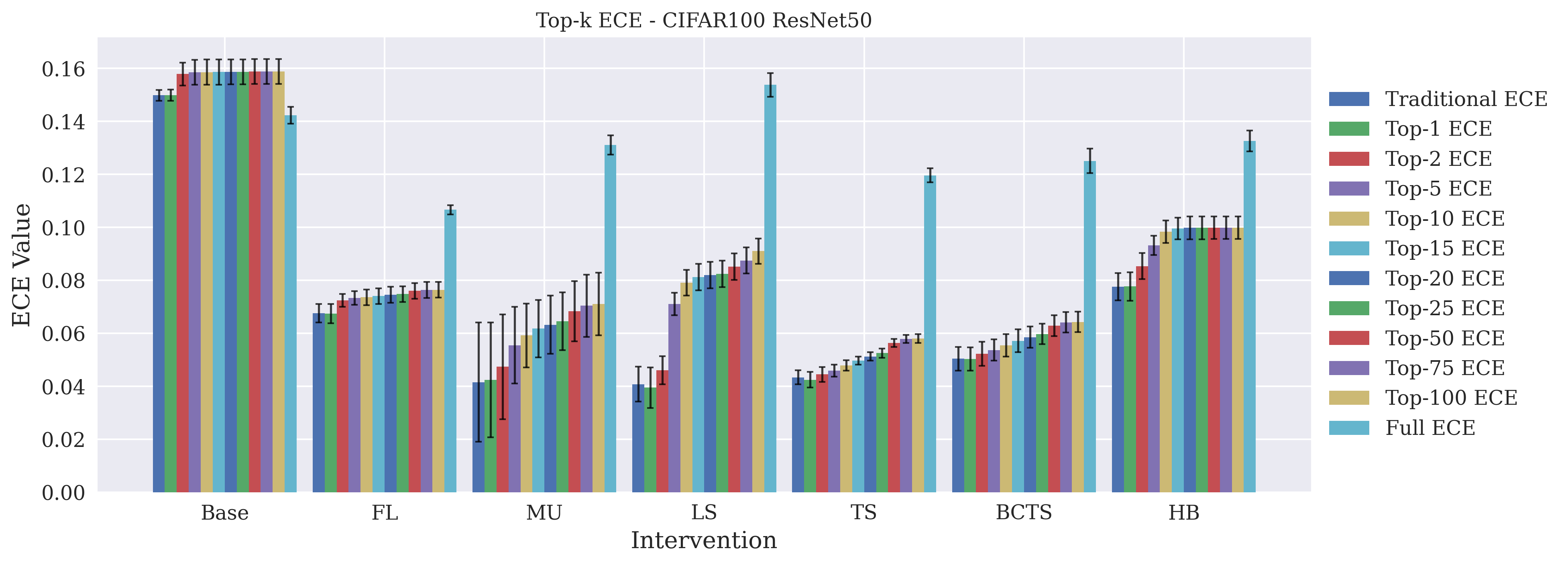}
    \caption{Top-$k$ ECE (All Interventions, ResNet50, CIFAR-100)}
    \label{fig:TopKECE}
\end{figure*}

\begin{figure}
    \centering
    \includegraphics[width=0.65\linewidth]{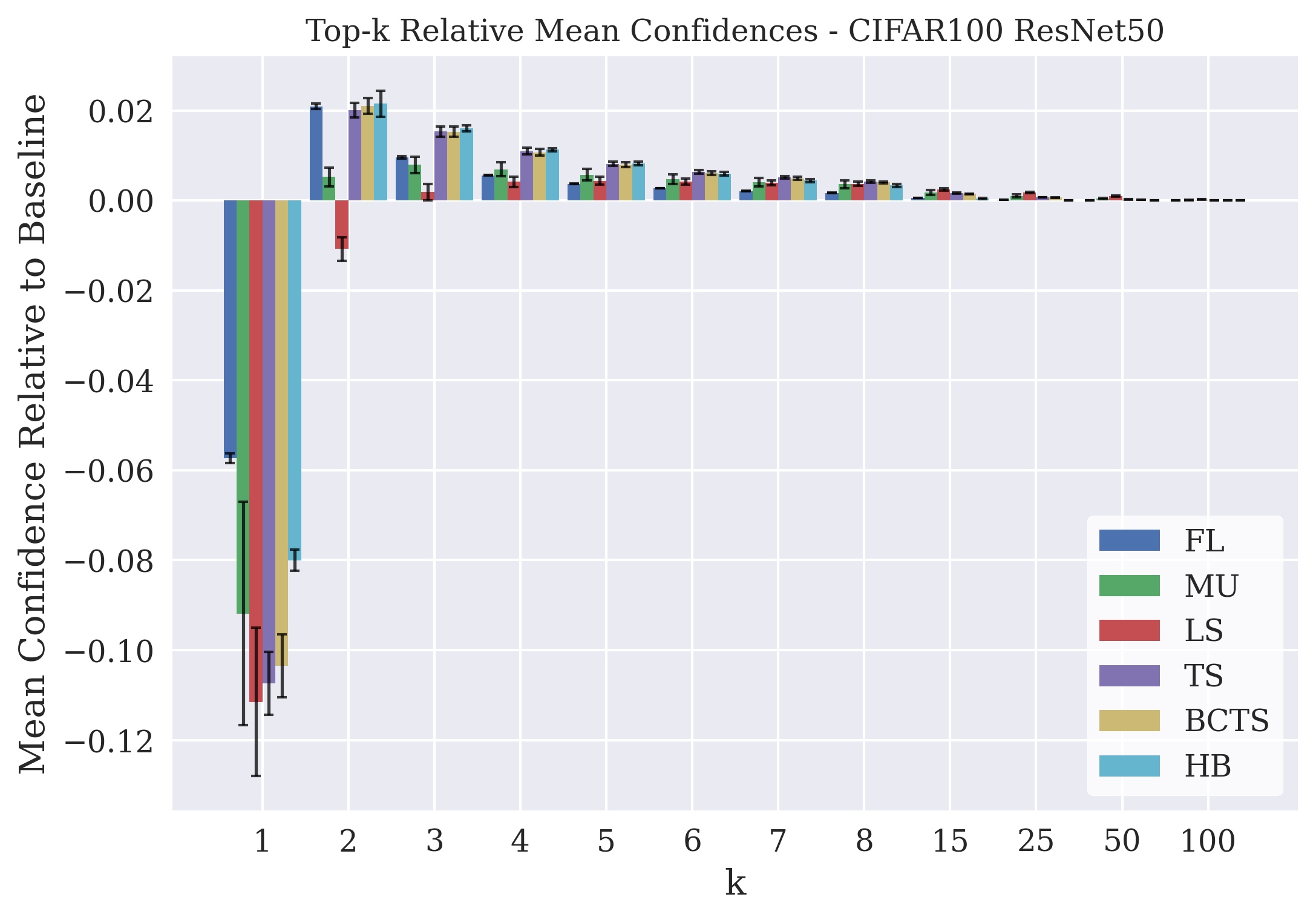}
    \caption{Mean Top-$k$ confidences, relative to baseline (All Interventions, ResNet50, CIFAR-100)}
    \label{fig:TopKConf}
\end{figure}

We perform the same experiment as in Section~\ref{sec:topKResults} of the main paper, but using the CIFAR-100~\citep{krizhevsky2009learning} data set. Similar to the visualizations in the main paper, Figure~\ref{fig:TopKECE} shows the top-$k$ ECE values for all interventions, and Figure~\ref{fig:TopKConf} shows the mean confidence values for the $k$-th highest-valued output class for all interventions. The biggest difference between these results and those obtained by evaluating the models on ImageNet is the performance of Focal Loss, as we did not have an issue finding a suitable hyperparameter setting for the intervention in this case. All interventions reduce the top-$k$ ECE over the baseline, however, each intervention trends differently as $k$ increases.
For example, focal loss has a higher top-1 ECE than label smoothing, but a lower top-100 ECE. Additionally, the improvement in Full ECE shown in Figure~\ref{fig:TopKECE}, does not correlate perfectly with improvement in Traditional ECE for all techniques (note label smoothing actually increases the full ECE relative to the baseline). As discussed in the main work, based on Figure~\ref{fig:TopKConf}, this phenomenon can likely be attributed to the fact that many of the techniques place more probability mass on less probably classes than the baseline, even though the correct class is rarely one of them.

\subsubsection{Grouping Lens and Label Selection}
Similar to the experiment in Section~\ref{sec:groupResults} of the main paper, we perform a grouping experiment this time on the CBIS-DDSM data set.
This data set consists of $224 \times 224$ patches extracted from mammography imagery.
Each patch is an instance in our experiment labeled as either ``benign mass'', ``benign calcification'', ``malignant mass'', ``malignant calcification'', or ``background''.
For groups in our experiments, we grouped the two ``malignant'' labels to form one group, the two ``benign'' labels to form a second, and used the ``background'' images as a third.
This induces a classification problem that tries to discriminate ``malignant'', ``benign'', and ``background'' patches.

\begin{figure*}
    \centering
    \includegraphics[width=\linewidth]{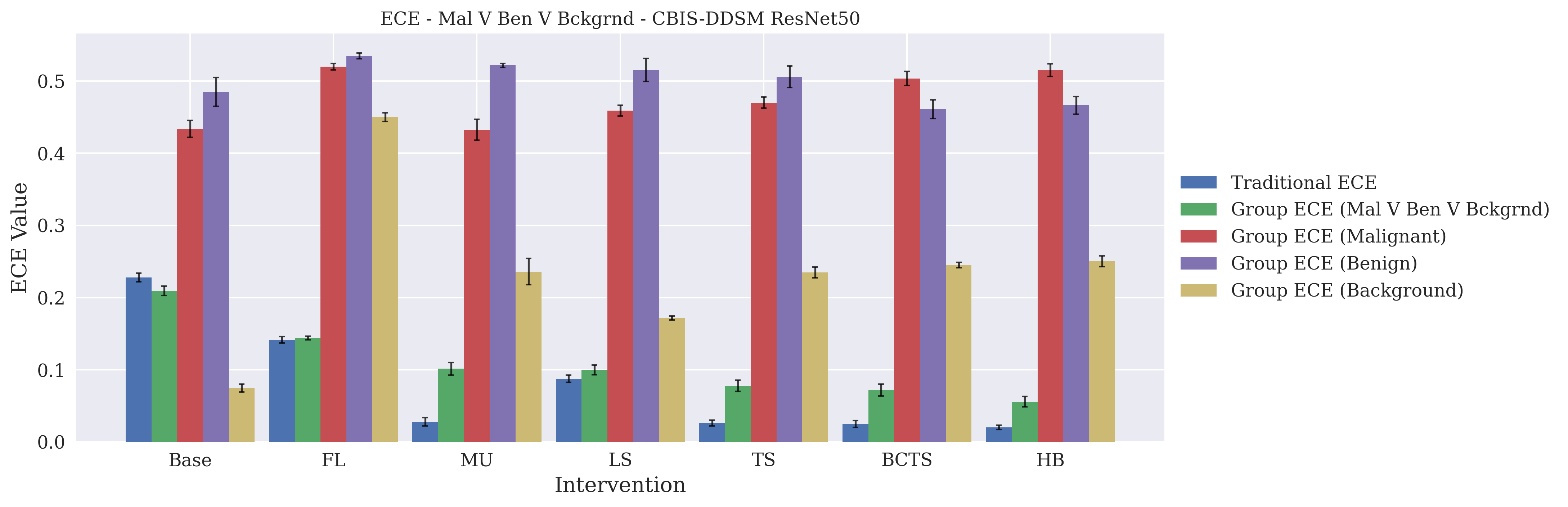}
    \caption{Group-wise ECEs, (All Interventions, ResNet50, CBIS-DDSM)}
    \label{fig:GroupingECECBIS}
\end{figure*}

Figure~\ref{fig:GroupingECECBIS} shows a number of ECE measures for all interventions on the CBIS-DDSM data set.
The green bar represents the full ECE after the aforementioned grouping is applied.
We can see that all interventions are able to reduce ECE for this induced problem as well as Traditional ECE (blue bar).
The remaining bars represent the same grouped ECE but after a group selection operator is applied for each group.
The main result here is that while all interventions are able to reduce the full group ECE over the baseline, most actually have higher group ECE when each group is isolated.
This example highlights that when adaptive binning is used separately for different ECE metrics, then the results between different ECE measures becomes difficult to directly relate to each other.
In this case, because different instances were used to find the adaptive bins for the full group and the group conditional ECEs, their realizations become decoupled.

\begin{figure}
    \centering
    \includegraphics[width=0.7\linewidth]{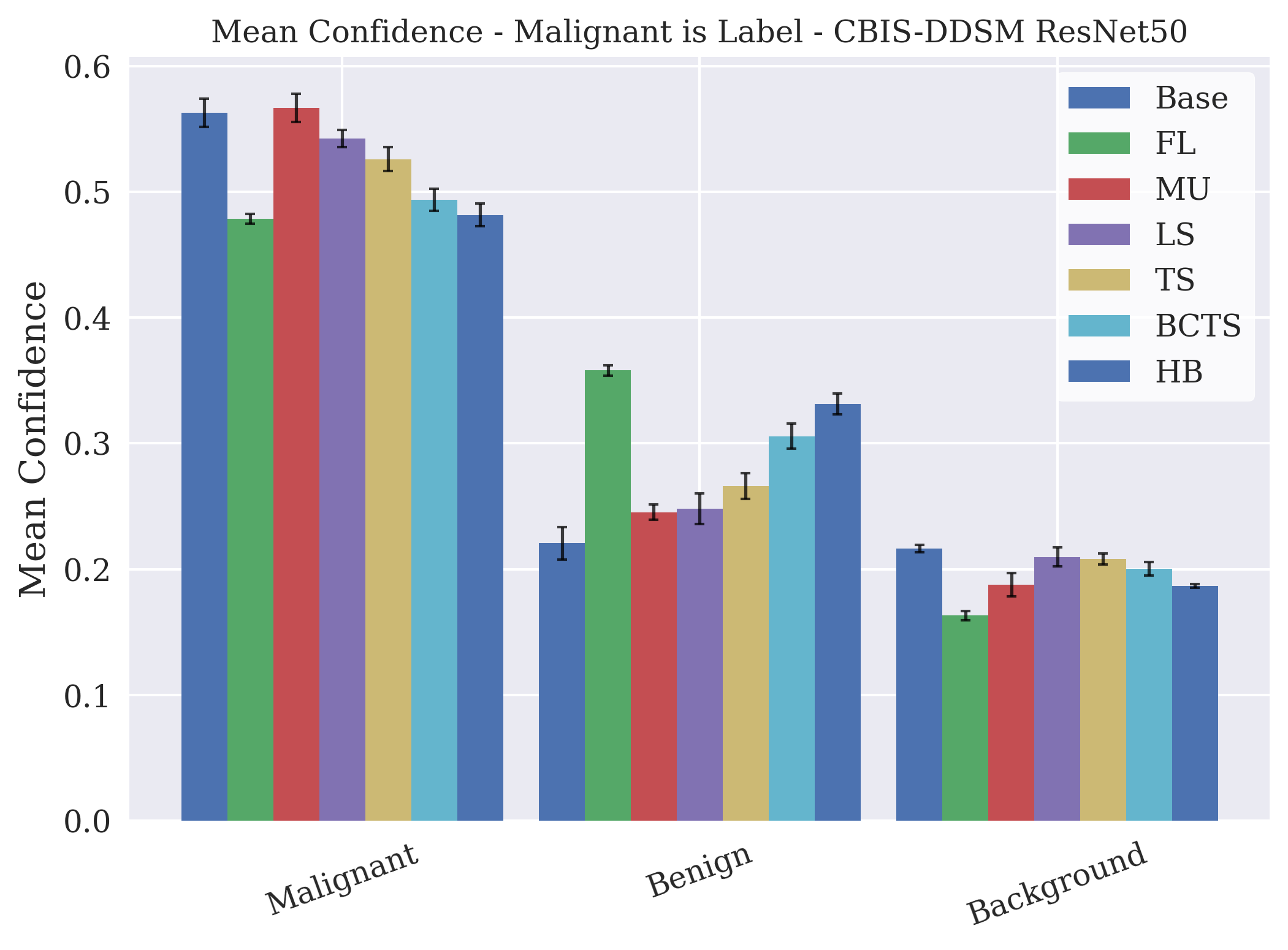}
    \caption{Mean group-wise confidence, conditioned on ``malignant'' group (All Interventions, ResNet50, CBIS-DDSM)}
    \label{fig:GroupingConfCBIS}
\end{figure}

Figure~\ref{fig:GroupingConfCBIS} shows the mean confidence values of all models for each of the three groups after the malignant conditional operator is applied, meaning only instances with ``malignant'' labels are evaluated.
Here we see a more detailed picture of calibration error than by just looking at Figure~\ref{fig:GroupingECECBIS}.
For instance, Focal Loss and BCTS incur very similar malignant conditional group ECE, but Focal Loss tends to put more probability mass on the benign class than the background class.
Such a difference between how error is incurred can have important practical ramifications.

\subsubsection{Inter-Interval Distance and Output Selection} 
Finally, we performed experiments similar to those in Section~\ref{sec:interIntervalResults} in the main paper to show another use case of inter-interval distance and output selection, this time using the COVID-CXR-2 data set.
This data set contains chest x-ray images labeled either as COVID positive or COVID negative.
Here we perform the same output selection operators and inter-interval distance functions to create ``very high confidence'' ([0.9,1.0]) and ``not very high confidence'' ([0.0,0.8]) intervals.
Such intervals can be used to alert clinicians when a patient is very likely to be COVID positive.

\begin{figure*}
    \centering
    \includegraphics[width=\linewidth]{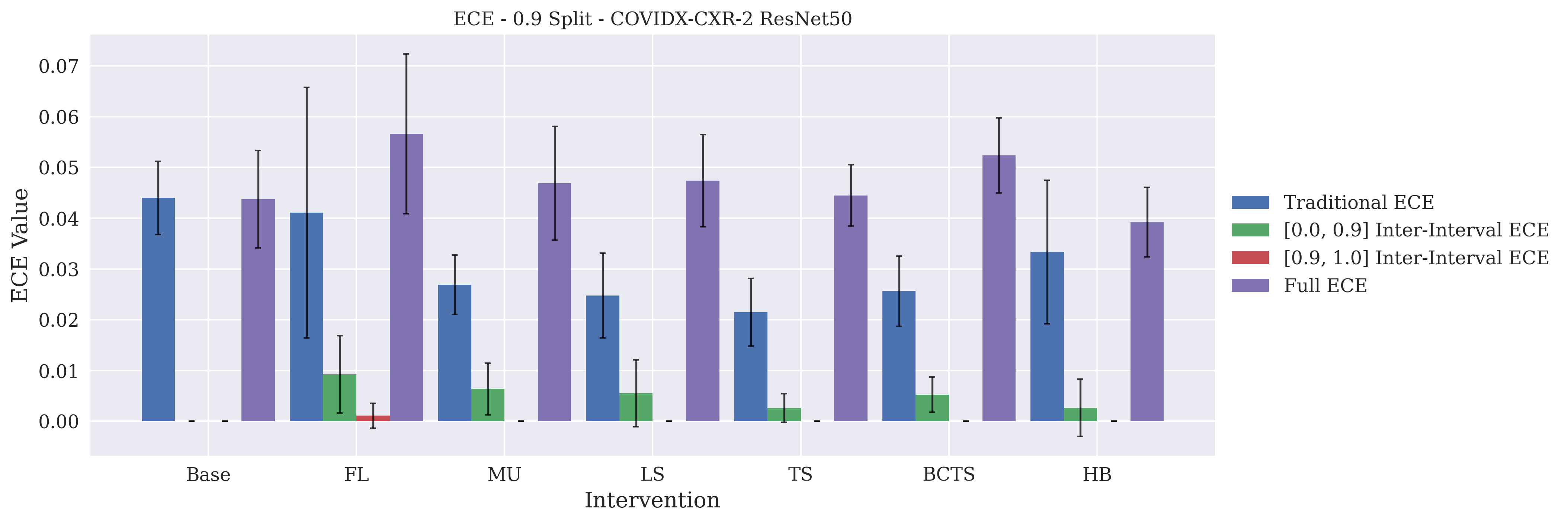}
    \caption{Interval-wise ECEs (All Interventions, ResNet50, COVIDX-CXR2)}
    \label{fig:IntervalECECOVID}
\end{figure*}

\begin{figure}
  \centering
    \includegraphics[width=0.8\linewidth]{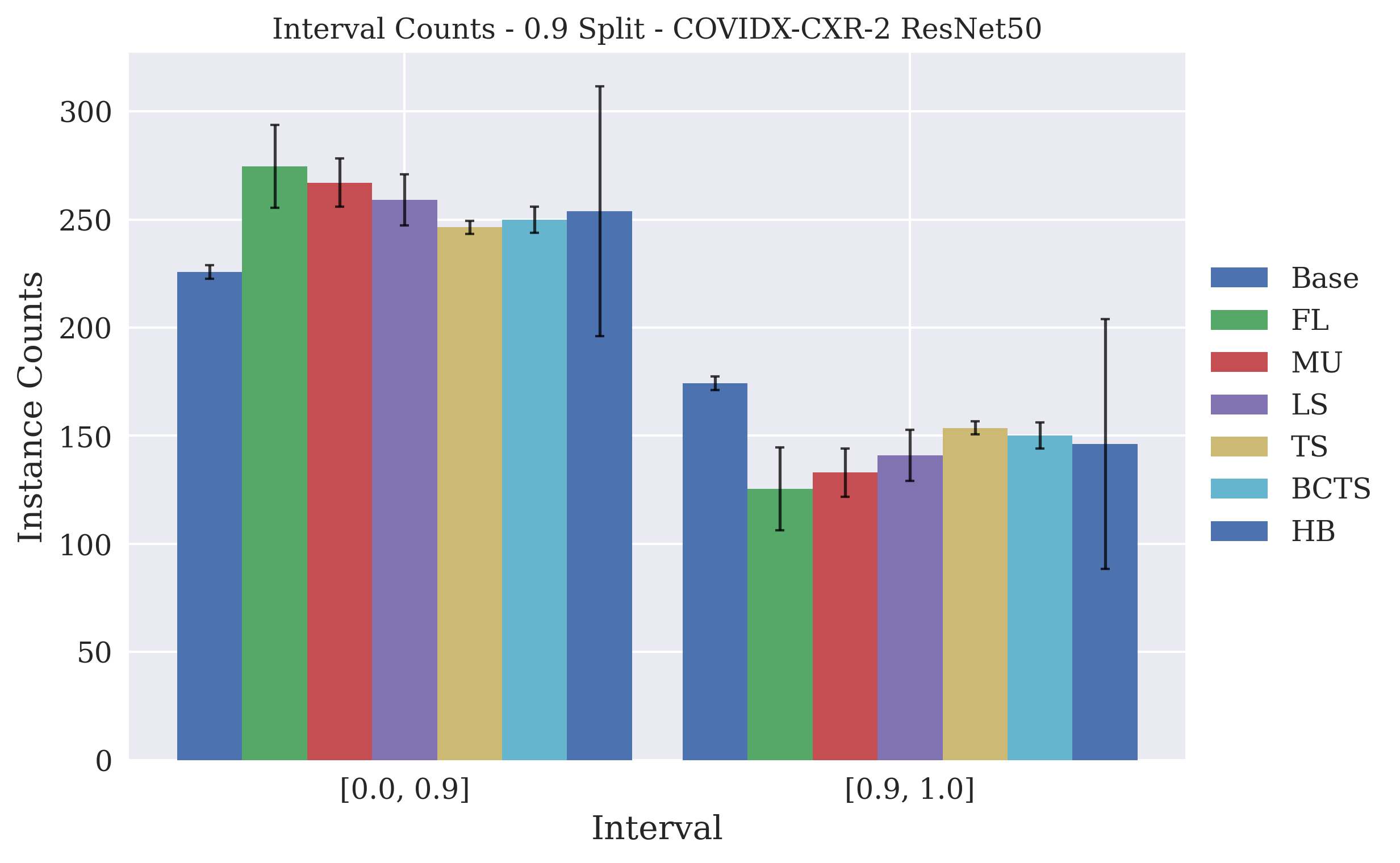}
    \caption{Test instances per interval (All Interventions, ResNet50, COVIDX-CXR2)}
  \label{fig:IntervalCountCOVID}
\end{figure}

Figure~\ref{fig:IntervalECECOVID} shows the Traditional ECE, Full ECE, and the intra-interval errors from both intervals created using output conditional operators and inter-interval distance, similar to what is done in the main paper.
The baseline model has zero error for both of the inter-interval errors, while all interventions incur a non-zero amount of error.
Figure~\ref{fig:IntervalCountCOVID} shows that all interventions move some points from the ``very high confidence'' interval to the ``not very high confidence'' interval.
This indicates that many of the instances moved were positive instances, making the mean labels in the ``not very high confidence'' interval shift to the ``very high confidence interval'' causing error.
This shows an example where the baseline model seemingly has high calibration error compared to interventions according to Traditional ECE, but actually is well suited to reflect a particular Likert categorization of confidence.

\clearpage


\begin{table*}[h]
\small
\vspace{-2em}


    \centering
    \begin{tabular}{|l|l|l|l|}
        \hline
        Model & ResNet50 & MobileNetV2 & DenseNet100 \\
        \hline
        Base Model & 0.739 ± 0.0052 & 0.734 ± 0.0022 & 0.791 ± 0.0038 \\
        Focal Loss & \textbf{0.764 ± 0.0035} & 0.728 ± 0.0021 & 0.795 ± 0.0053 \\
        Mix-Up & 0.757 ± 0.0037 & \textbf{0.746 ± 0.0036} & \textbf{0.802 ± 0.0038} \\
        Label Smoothing & 0.744 ± 0.0115 & 0.735 ± 0.0023 & 0.791 ± 0.0021 \\
        Temp. Scaling & 0.739 ± 0.0052 & 0.734 ± 0.0022 & 0.791 ± 0.0038 \\
        Bias-Corrected TS & 0.735 ± 0.0082 & 0.731 ± 0.0021 & 0.792 ± 0.0020 \\
        Hist. Binning & 0.731 ± 0.0056 & 0.728 ± 0.0032 & 0.785 ± 0.0033 \\
        \hline
    \end{tabular}
    \caption{Top-1 Classification Accuracy, CIFAR-100 Data Set}

\bigskip
    \centering
    \begin{tabular}{|l|l|l|l|}
        \hline
        Model & ResNet50 & MobileNetV2  & DenseNet201 \\
        \hline
        Base Model & 0.761 ± 0.0013 & 0.716 & 0.774 \\
        Focal Loss & 0.742 ± 0.0002 & 0.682 & 0.757 \\
        Mix-Up & \textbf{0.768 ± 0.0005} & 0.719 & \textbf{0.782} \\
        Label Smoothing & 0.766 ± 0.0016 & 0.717 & 0.778 \\
        Temp. Scaling & 0.761 ± 0.0013 & 0.716 & 0.774 \\
        Bias-Corrected TS & 0.765 ± 0.0002 & \textbf{0.721} & 0.779 \\
        Hist. Binning & 0.760 ± 0.0008 & 0.716 & 0.775 \\
        \hline
    \end{tabular}
    \caption{Top-1 Accuracy, ImageNet (ResNet50/MobileNetV2/DenseNet201 Trials: 2/1/1)}

\bigskip
    \centering
    \begin{tabular}{|l|l|l|l|}
        \hline
        Model & ResNet50 & MobileNetV2 & DenseNet201 \\
        \hline
        Base Model & 0.694 ± 0.0076 & 0.681 ± 0.0054 & 0.692 ± 0.0087 \\
        Focal Loss & 0.663 ± 0.0050 & 0.652 ± 0.0038 & 0.668 ± 0.0074 \\
        Mix-Up & \textbf{0.707 ± 0.0048} & \textbf{0.697 ± 0.0046} & \textbf{0.714 ± 0.0035} \\
        Label Smoothing & 0.689 ± 0.0051 & 0.684 ± 0.0068 & 0.691 ± 0.0069 \\
        Temp. Scaling & 0.694 ± 0.0076 & 0.681 ± 0.0054 & 0.692 ± 0.0087 \\
        Bias-Corrected TS & 0.696 ± 0.0063 & 0.682 ± 0.0035 & 0.697 ± 0.0081 \\
        Hist. Binning & 0.683 ± 0.0076 & 0.665 ± 0.0083 & 0.679 ± 0.0073 \\
        \hline
    \end{tabular}
    \caption{Top-1 Classification Accuracy, CBIS-DDSM Data Set}

\bigskip
    \centering
        \begin{tabular}{|l|l|l|l|}
                \hline
                Model & ResNet50 & MobileNetV2 & DenseNet201 \\
                \hline
                Base Model & 0.844 ± 0.0049 & 0.811 ± 0.0052 & 0.844 ± 0.0088 \\
                Focal Loss & 0.842 ± 0.0050 & 0.806 ± 0.0063 & 0.840 ± 0.0074 \\
                Mix-Up & \textbf{0.858 ± 0.0051} & 0.803 ± 0.0067 & \textbf{0.853 ± 0.0102} \\
                Label Smoothing & \textbf{0.858 ± 0.0011} & \textbf{0.818 ± 0.0044} & 0.848 ± 0.0072 \\
                Temp. Scaling & 0.844 ± 0.0049 & 0.811 ± 0.0052 & 0.844 ± 0.0088 \\
                Bias-Corrected TS & 0.848 ± 0.0039 & 0.813 ± 0.0071 & 0.849 ± 0.0090 \\
                Hist. Binning & 0.815 ± 0.0046 & 0.781 ± 0.0069 & 0.814 ± 0.0071 \\
                \hline
        \end{tabular}
        \caption{Top-1 Classification Accuracy, ETHEC Data Set}
        
    \bigskip
    \centering
        \begin{tabular}{|l|l|l|l|}
                \hline
                Model & ResNet50 & MobileNetV2 & DenseNet201 \\
                \hline
                Base Model & \textbf{0.945 ± 0.0060} & 0.949 ± 0.0087 & 0.962 ± 0.0098 \\
                Focal Loss & 0.930 ± 0.0133 & 0.950 ± 0.0114 & 0.951 ± 0.0063 \\
                Mix-Up & \textbf{0.945 ± 0.0129} & \textbf{0.956 ± 0.0042} & \textbf{0.972 ± 0.0068} \\
                Label Smoothing & 0.939 ± 0.0110 & 0.948 ± 0.0123 & 0.967 ± 0.0054 \\
                Temp. Scaling & \textbf{0.945 ± 0.0060} & 0.949 ± 0.0087 & 0.962 ± 0.0098 \\
                Bias-Corrected TS & 0.939 ± 0.0055 & 0.943 ± 0.0042 & 0.953 ± 0.0102 \\
                Hist. Binning & \textbf{0.945 ± 0.0054} & 0.952 ± 0.0113 & 0.954 ± 0.0193 \\
                \hline
        \end{tabular}
    \caption{Top-1 Classification Accuracy, COVIDX-CXR-2 Data Set}
        
    \bigskip
    \centering
        \begin{tabular}{|l|l|l|l|}
                \hline
                Model & ResNet50 & MobileNetV2 & DenseNet201 \\
                \hline
                Base Model & 0.825 ± 0.0023 & 0.813 ± 0.0060 & 0.824 ± 0.0042 \\
                Focal Loss & 0.823 ± 0.0025 & 0.814 ± 0.0035 & 0.828 ± 0.0033 \\
                Mix-Up & \textbf{0.828 ± 0.0043} & \textbf{0.831 ± 0.0045} & \textbf{0.830 ± 0.0048} \\
                Label Smoothing & 0.819 ± 0.0033 & 0.811 ± 0.0054 & 0.825 ± 0.0026 \\
                Temp. Scaling & 0.825 ± 0.0023 & 0.813 ± 0.0060 & 0.824 ± 0.0042 \\
                Bias-Corrected TS & 0.826 ± 0.0023 & 0.814 ± 0.0066 & 0.825 ± 0.0025 \\
                Hist. Binning & 0.806 ± 0.0027 & 0.801 ± 0.0055 & 0.817 ± 0.0057 \\
                \hline
        \end{tabular}
        \caption{Top-1 Classification Accuracy, CrisisMMD Data Set}
\end{table*}


%
\clearpage
\clearpage
%

%
%
%
%

\end{document}

%% file: math_commands.tex
\usepackage{amsmath,amsfonts,bm}

\def\eqref#1{equation~\ref{#1}}

\def\1{\bm{1}}

\DeclareMathAlphabet{\mathsfit}{\encodingdefault}{\sfdefault}{m}{sl}
\SetMathAlphabet{\mathsfit}{bold}{\encodingdefault}{\sfdefault}{bx}{n}

\DeclareMathOperator*{\argmax}{arg\,max}